\documentclass[11pt]{article}

\usepackage[final]{acl}

\usepackage{times}
\usepackage{latexsym}

\usepackage[T1]{fontenc}

\usepackage[utf8]{inputenc}

\usepackage{microtype}

\usepackage{inconsolata}

\usepackage{graphicx}

\usepackage{hyperref}
\usepackage{url}
\usepackage{booktabs}
\usepackage{amsfonts}
\usepackage{nicefrac}
\usepackage{xcolor}
\usepackage{amsmath}
\usepackage{amssymb}
\usepackage{amsthm}
\usepackage{listings}
\usepackage{hwemoji}

\lstdefinestyle{latexstyle}{
    language=[LaTeX]TeX,
    backgroundcolor=\color{white},
    basicstyle=\ttfamily\small,
    keywordstyle=\color{blue},
    commentstyle=\color{gray},
    stringstyle=\color{red!70!black},
    numbers=left,
    numberstyle=\tiny\color{gray},
    numbersep=6pt,
    xleftmargin=14pt, 
    frame=single,
    rulecolor=\color{black!20},
    breaklines=true,
    showstringspaces=false
}

\usepackage{caption}
\usepackage{subcaption}
\usepackage{tikz}
\usetikzlibrary{trees}

\newtheorem{theorem}{Theorem}[section]
\theoremstyle{definition}
\newtheorem{remark}[theorem]{Remark}

\title{\texttt{.tmu}: A Low-Entropy Tree-Structured Representation for LLM-Assisted Scientific Writing}

\author{
Tianyou Liu$^{\dagger,1}$,
Ziqiang Li$^{\dagger,2}$,
Xurui Liu$^{3}$,
Yu Wu$^{4}$,
Yansong Li$^{\ast,5}$\\[0.5em]
$^{1}$Southern University of Science and Technology,
$^{2}$Alibaba,\\
$^{3}$Tsinghua University,
$^{4}$Rutgers University,
$^{5}$Eastern Institute of Technology, Ningbo
}

\newcommand{\dagfootnote}{%
  {\let\thefootnote\relax
  \footnotetext[0]{%
    \begin{tabular}{@{}l@{\hspace{0.25em}}l@{}}
      $^\dagger$ & Equal Contribution.\\
      $^\ast$    & Corresponding author.
      \href{mailto:ysli@eitech.edu.cn}{\texttt{ysli@eitech.edu.cn}}
    \end{tabular}
  }}%
}

\begin{document}

\maketitle
\dagfootnote

\begin{abstract}
As large language models (LLMs) increasingly assist scientific writing, the limitations and token costs of generating {\TeX} become increasingly visible. This paper analyzes {\TeX}'s architectural mismatch with LLM workflows, stemming from its lack of an explicit structural representation, to illustrate its limitations on generated semantics and error localization. As an alternative, we introduce \texttt{.tmu}, a low-entropy tree-structured representation. With its efficient data structure and clear contextual boundaries, \texttt{.tmu} outperforms \texttt{.tex} in the above aspects. Experiments across four LLMs provide evidence for this claim in most evaluated settings. Furthermore, we show that due to its lower information entropy, fine-tuning LLMs on \texttt{.tmu} achieves approximately 43\% lower final training loss than on \texttt{.tex}. Our work provides a more scalable and LLM-friendly data representation for LLM-assisted scientific writing.
\end{abstract}

\section{Introduction}
\label{sec:introduction}

LLM-assisted scientific writing is moving from isolated chat interactions to production workflows: agents inspect manuscripts, draft equations and proofs, merge revisions, and repair build failures in editors, CI systems, and conversion pipelines. Most such workflows still operate over {\LaTeX} source. {\LaTeX} remains excellent for typography, mathematical notation, and long-term compatibility, but its source format is optimized for typesetting rather than machine action. For LLM systems, the bottleneck is therefore representational: tools must recover document structure from a macro program that was not designed to expose it.

Three practical failure modes follow. First, {\LaTeX} exposes no explicit document tree, so agents spend tokens rediscovering sections, labels, equations, and scope. Second, structure and rendering instructions are coupled, making static analysis unreliable without executing the document. Third, errors are often non-local: a small syntactic defect can surface many lines later and send an agent into repeated compile-and-repair loops. In industry settings, these failures translate directly into latency, tool calls, compute cost, and lower reliability (we detail the mechanisms in Section~\ref{sec:background}).

We study \texttt{.tmu}, a low-entropy tree-structured representation from the GNU {\TeX}macs lineage~\cite{van_der_hoeven_gnu_2001}, as an alternative substrate for LLM-assisted scientific writing. In \texttt{.tmu}, formulas, citations, sections, and layout objects are represented as addressable tree nodes. This makes local edits and reference updates explicit and reduces source-level ambiguity: many equivalent {\LaTeX} encodings collapse into a more canonical tree. The same representation also supports WYSIWYG authoring for humans, but our claim here is machine-facing: structured, low-entropy source should be easier for LLM systems to locate, merge, repair, and model. In practice, \texttt{.tmu} already underlies an existing scientific-writing system supporting document import, structured editing, node-level LLM assistance, syntax checking, and local repair.

Our contributions are as follows:
\vspace{-0.35em}
\begin{itemize}
    \setlength{\itemsep}{0.15em}
    \setlength{\parsep}{0pt}
    \setlength{\parskip}{0pt}
    \item We diagnose three representational frictions of {\LaTeX} for LLM authoring---no explicit AST, structure--typesetting coupling, and non-local error reporting---and tie each to a concrete LLM failure mode.
    \item We present a low-entropy, tree-structured \texttt{.tmu} representation (on the {\TeX}macs lineage) with explicit semantic nodes and clear contextual boundaries.
    \item We empirically evaluate \texttt{.tmu} on structure location, style-aware document merging, debugging of ill-formed documents, and formula-completion fine-tuning, showing gains in most evaluated settings (Section~\ref{sec:experiments}).
\end{itemize}

\section{Background: Why {\TeX} Limits LLM Workflows}
\label{sec:background}

This section details the frictions summarized in Section~\ref{sec:introduction}.

\subsection{Introduction of {\TeX}}

For decades, {\LaTeX} has been the standard for scientific writing~\cite{lamport1994document}. As an unstructured macro language~\cite{van_der_hoeven_jolly_2020}, however, it presents structural limitations for modern LLM-driven workflows along the axis of machine-tractable document structure \emph{(we defer a brief history of {\TeX} to Appendix~\ref{app:intro})}.

\subsection{Principles of {\TeX} Compilation}

{\TeX} uses a batch-processing model: it reads source code from beginning to end, expanding macros step by step without building an explicit document tree (Abstract Syntax Tree) in memory~\cite{van_der_hoeven_gnu_2001}. Lacking internal state, {\LaTeX} must resolve cross-references and bibliographies through multiple compilation passes, writing temporary data into intermediate files (\texttt{.aux}, \texttt{.toc}, \texttt{.bbl}) and reading them back on the next run \emph{(detailed in Appendix~\ref{app:pr})}. This multi-pass design is inherently slow and opaque, precluding the sub-second, real-time feedback required by modern editors and AI agents~\cite{madje2023typst}.

\subsection{Architectural Defects in {\TeX}}

{\TeX} processes documents as a linear sequence of commands in which structural semantics and typesetting decisions are deeply coupled within a single execution path. Consequently, AI models and static analysis tools cannot reliably extract the document's structure without executing the code \emph{(see Appendix~\ref{app:arch} for a deep analysis)}. This separation problem is fundamental, not incidental: Frank Mittelbach, the technical lead for {\LaTeX}, candidly conceded in an interview~\cite{interview2021} that
\begin{quote}
the separation between local and global in {\LaTeX} seems to be hard. [\ldots] First of all, it is right now hard in {\TeX}\ldots
\end{quote}

The same coupling makes error feedback delayed and ambiguous: the reported location of an error is rarely where the mistake actually happened, and a subtle local defect can irreversibly abort the entire compilation. Listing~\ref{lst:unclosed-frac} shows a typical case---the second argument of \verb|\frac| is missing its closing brace, yet {\TeX} keeps consuming subsequent tokens and only reports a \texttt{Runaway argument?} error at \verb|\end{equation}|, far from the true cause.

\begin{lstlisting}[caption={Unclosed brace: error reported at \texttt{\textbackslash end\{equation\}}, not at its source},label={lst:unclosed-frac}]
\begin{equation}
x=\frac{-b\pm \sqrt{b^{2}-4ac}}{2a
\end{equation}
\end{lstlisting}

Because such logs provide little structural guidance, models are prone to prolonged, often non-terminating debugging loops and hallucinations~\cite{jain2026bibby}. This mechanism constrains LLM-driven writing.

\subsection{UX \& Ecosystem Constraints}

As detailed in Appendix~\ref{app:con}, {\LaTeX}'s ecosystem is too heavy and fragmented: a full TeX Live distribution exceeds 6GB, hindering deployment in cloud environments or sandboxes, while different engines (pdfLaTeX, XeLaTeX, LuaLaTeX) handle fonts and Unicode differently, yielding inconsistent results. While {\LaTeX} retains clear technical strengths in typographic quality and mathematical expressivity, its continued dominance is also sustained by established user habits and publisher requirements~\cite{van_der_hoeven_jolly_2020}; for LLM-driven workflows specifically, these factors leave its structural limitations unaddressed. This motivates a more scalable, LLM-friendly data representation.

\paragraph{Related Work}
The friction between \LaTeX{} and LLM-driven workflows is well-documented across three critical dimensions.

First, \textbf{generation and execution failures}. Benchmarks reveal that LLMs severely struggle with \LaTeX{}'s verbose syntax and long-range dependencies, showing low accuracy~\cite{kale2025texpert} and high token inefficiency in multimodal tasks~\cite{ling_table2latex_2025, xia2024docgenome}. This causes systemic bottlenecks in automated science pipelines~\cite{lu2024aiscientist} and ``execution illusions'' of uncompilable code~\cite{lyn2025translatex}. Thus, augmenting compiler logs with live ASTs is necessary for reliable AI assistance~\cite{jain2026bibby}.

Second, \textbf{training and representation overhead}. Upstream, processing heterogeneous \LaTeX{} incurs massive engineering overhead~\cite{paster2023openwebmath} and prohibitive token costs~\cite{lin2024accurate}. \LaTeX{}'s syntactic redundancy reduces predictability and inflates training costs~\cite{taylor2022galactica}. Conversely, explicitly modeling markup hierarchies drastically improves document understanding~\cite{li2022markuplm} and AST-guided code generation~\cite{zeng2026treediffastguidedcodegeneration}.

Third, \textbf{system architecture limitations}. To address batch compilation delays, existing solutions propose incremental compilers~\cite{madje2023typst, haug2022typst} or GUI overlays~\cite{gobert2022ilatex}, but still separate authoring from visual output. Meanwhile, studies highlight \LaTeX{}'s real-world fragility~\cite{tan2024inconsistencies}, user inefficiency~\cite{knauff2014efficiency}, and extreme runtime costs when used as a computing substrate~\cite{gardner2025neuralatex}. We build on the tree-based WYSIWYG lineage of \TeX{}macs~\cite{van_der_hoeven_gnu_2001}, but our contribution is distinct: whereas \TeX{}macs is prior art designed for human editing, we treat its \texttt{.tmu} format as a low-entropy, tree-structured \emph{representation for LLMs} and contribute the first evaluation of such a structured representation on LLM generation and fine-tuning. In doing so, we provide evidence that a tree-structured format offers advantages over \LaTeX{} along both axes the literature above identifies as broken---structural transparency for generation and predictability for training. \textit{(A comprehensive review is deferred to Appendix~\ref{app:rew}.)}

\section{The \texttt{.tmu} Representation: A Low-Entropy Tree Structure}
\label{sec:mogan}

While {\TeX} represents formulas through layout-oriented markup in the form of typesetting instructions, the \texttt{.tmu} format adopts a tree-based, functional representation for both mathematical formulas and document structure. This tree-structured representation shows advantages over the formats underlying other editors in Table~\ref{tab:mogan-vs-other}. \emph{(A detailed introduction to the \texttt{.tmu} representation and a comparison are provided in Appendix~\ref{app:mvse}.)} The following discussion evaluates this representation with specific examples.

The central property of our structured representation is that it is \emph{low-entropy}, by which we mean reduced source-encoding multiplicity. In \LaTeX{}, a single rendered output can be produced by many equivalent source strings: \verb|x^2| and \verb|x^{2}| render identically, \verb|\frac{a}{b}| and \verb|{a \over b}| denote the same fraction, and arbitrary whitespace can be inserted almost anywhere without changing the result. In the \texttt{.tmu} tree, by contrast, each rendered structure corresponds to an essentially canonical source, so this one-to-many redundancy is largely eliminated. Fewer equivalent encodings mean lower next-token uncertainty and therefore lower information entropy. This low-entropy property is the property that the remainder of this section instantiates through concrete examples.

The same low-entropy representation benefits both human authors and LLMs. Because the tree directly captures the rendered document, it supports a What You See Is What You Get (WYSIWYG) editing interface in which humans edit the rendered document in place, rather than following \LaTeX{}'s edit-source-then-compile loop. In this paper, however, we make no quantitative claim about human input speed; our measured contribution is the machine-facing payoff, where the lower-entropy structure is easier for LLMs to locate, merge, repair, and model.

\subsection{Tree-Structured Formulas and Document Structure}
\label{sec:tree-struc-on-mogan}

Unlike \LaTeX{}'s compilation model, which centers on linear text and macro expansion, \TeX{}macs features an explicit tree-based document structure designed from the outset. Document components---such as chapters, formulas, citations, and typesetting elements---exist as structured nodes, rather than being implicitly embedded within macro calls or token streams. This architectural distinction directly impacts citation updates, compilation efficiency, and the interactive user experience.

A fraction serves as a primary example of this difference. In \TeX{}, a fraction is represented as \verb|\frac{1}{2}|; internally, the compiler processes \verb|\frac|, \verb|{1}|, and \verb|{2}| serially. In the \texttt{.tmu} representation, however, the fraction is represented as the tree \verb|(frac "1" "2")|, written here in a compact S-expression notation (in the style of Lisp/Scheme). This forms a tree structure, as illustrated in Figure~\ref{fig:frac-tree}.

\begin{figure}[htbp]
    \vspace{-4mm}
    \centering
    \small
    \begin{tikzpicture}[
            level 1/.style={sibling distance=15mm, level distance=7mm},
            every node/.style={align=center}
        ]
        \node {\texttt{frac}}
        child {node {\texttt{"1"}}}
        child {node {\texttt{"2"}}};
    \end{tikzpicture}
    \caption{The Tree Structure of \texttt{.tmu} Formulas}
    \label{fig:frac-tree}
\end{figure}

As a further example, consider the mathematical expression in Equation~\eqref{eq:tree-struc}:

\begin{equation}
    \label{eq:tree-struc}
    \int_{a}^{b} f(x) \, \mathrm{d}x = \left[ F(x) \right] \Big|_{a}^{b}
\end{equation}

Its \texttt{.tmu} tree\footnote{On disk, a \texttt{.tmu} file stores this tree in an XML-like serialization; throughout this paper we display the same tree in the equivalent S-expression notation for readability, the two being interconvertible without information loss.} is shown in Listing~\ref{lst:scheme-integral}:
\begin{lstlisting}[caption={The \texttt{.tmu} tree of the integral, in S-expression notation},label={lst:scheme-integral},upquote=true]
> `(math (concat (big "int") (rsub "a") (rsup "b") "f" (around* "(" "x" ")") "<mathd>x=" (around* "<nobracket>" (around* "[" (concat "F" (around* "(" "x" ")")) "]") "|") (rsub "a") (rsup "b")))
\end{lstlisting}

Note that the segment $\left[ F(x) \right] |$ is managed via the tree structure illustrated in Figure~\ref{fig:tree-representation}. Consequently, even if a user omits a parenthesis or two, this omission does not negatively affect the rendering of the entire mathematical expression under the \texttt{.tmu} representation.

\begin{figure}[htbp]
    \vspace{-4mm}
    \centering
    \small
    \begin{tikzpicture}[
            level 1/.style={sibling distance=15mm, level distance=7mm},
            level 2/.style={sibling distance=12mm, level distance=7mm},
            level 3/.style={sibling distance=12mm, level distance=7mm},
            every node/.style={align=center}
        ]
        \node {\texttt{around*}}
        child {node {\texttt{"["}}}
        child {node {\texttt{concat}}
                child {node {\texttt{"F"}}}
                child {node {\texttt{around*}}
                        child {node {\texttt{"("}}}
                        child {node {\texttt{"x"}}}
                        child {node {\texttt{")"}}}
                    }
            }
        child {node {\texttt{"]"}}}
        child {node {\texttt{" | "}}};
    \end{tikzpicture}
    \caption{Tree representation of $ [F (x)] |$}
    \label{fig:tree-representation}
\end{figure}

Unlike \TeX{}, this tree structure exists at the rendering level, not the syntax level. This rendering-level tree structure offers two distinct advantages:

\begin{enumerate}
    \item \textbf{Local Scoping of Input Errors:}
          \begin{itemize}
              \item Errors do not cause the entire document to fail rendering (a major drawback of \LaTeX{}).
              \item Local edits do not trigger global layout instability (a major drawback of MS Word).
          \end{itemize}

    \item \textbf{Independent Subtree Layout:} Because the document is represented as a tree, independent subtrees can be laid out separately rather than strictly serially, which can reduce the work required to re-render after a local edit.
\end{enumerate}

The same locality extends beyond formulas: each line, image block, and structural element maps to an independent tree node, so local edits such as resizing an image need not disrupt surrounding layout.

\subsection{Functional Symbol Representation}

In \TeX{}, all mathematical symbols and fonts are represented as strings. However, in the \texttt{.tmu} representation, certain special symbols are defined via functions. For instance, the inner product $\langle \cdot \rangle$, as shown in Equation~\eqref{eq:braket}, is an anonymous function that automatically scales according to the symbols it contains. While this is similar to \verb|\langle \rangle| in \LaTeX{}, the difference lies in the fact that, as a complete function, the left and right brackets always render a unified pair rather than isolated characters. To obtain a single bracket, a structured editor operating on \texttt{.tmu} lets the user delete the undesired character from the rendered pair.

\begin{equation}
    \label{eq:braket}
    \left\langle \int \right\rangle \langle f \rangle
\end{equation}

\subsection{Dynamic System Optimization}

The \texttt{.tmu} representation encodes citation relationships as structural links directly embedded in the document tree. This mechanism avoids a full re-parsing of the document when updating. Furthermore, because complexity is dictated by what the document tree actually contains, an editor operating on \texttt{.tmu} can load functionality on demand rather than relying on a pre-configured feature set, so that system complexity is dictated by document content. This mechanism significantly reduces the burden for deploying cloud environments or sandboxes. \emph{(Further details on reference rendering and plugin loading are available in Appendix~\ref{app:ren} and Appendix~\ref{app:plug}.)}

\section{Numerical experiments}
\label{sec:experiments}

We compare \texttt{.tmu} with \texttt{.tex} on compiling/rendering speed, LLM task performance, and fine-tuning.

\paragraph{Setup.} All experiments derive from a 120-page arXiv document (\texttt{arXiv:2502.17655}) and run on an Intel Ultra 9 285H CPU, GeForce RTX 5080 GPU, and 32GB RAM. We evaluate four LLMs---Deepseek-v3.2, Claude-Sonnet-4.5, Gemini-3-pro, and ChatGPT-5.2---on structure locating (20 questions), style-aware file merging (5 tasks), and debugging (20 samples). Scores reward correctness while penalizing total token usage $T$ (input, thinking, output, and MCP tools), and timing benchmarks report the average of three trials per document. We refer to these scores as Cost-aware Task Utility. Incorrect outputs receive zero, while token cost discounts successful completions; a correct but over-budget response may also score zero, without being considered semantically equivalent to a hallucination.

\subsection{Benchmark on compiling/rendering time}

\begin{figure}[htbp]
    \centering
    \includegraphics[width=\columnwidth]{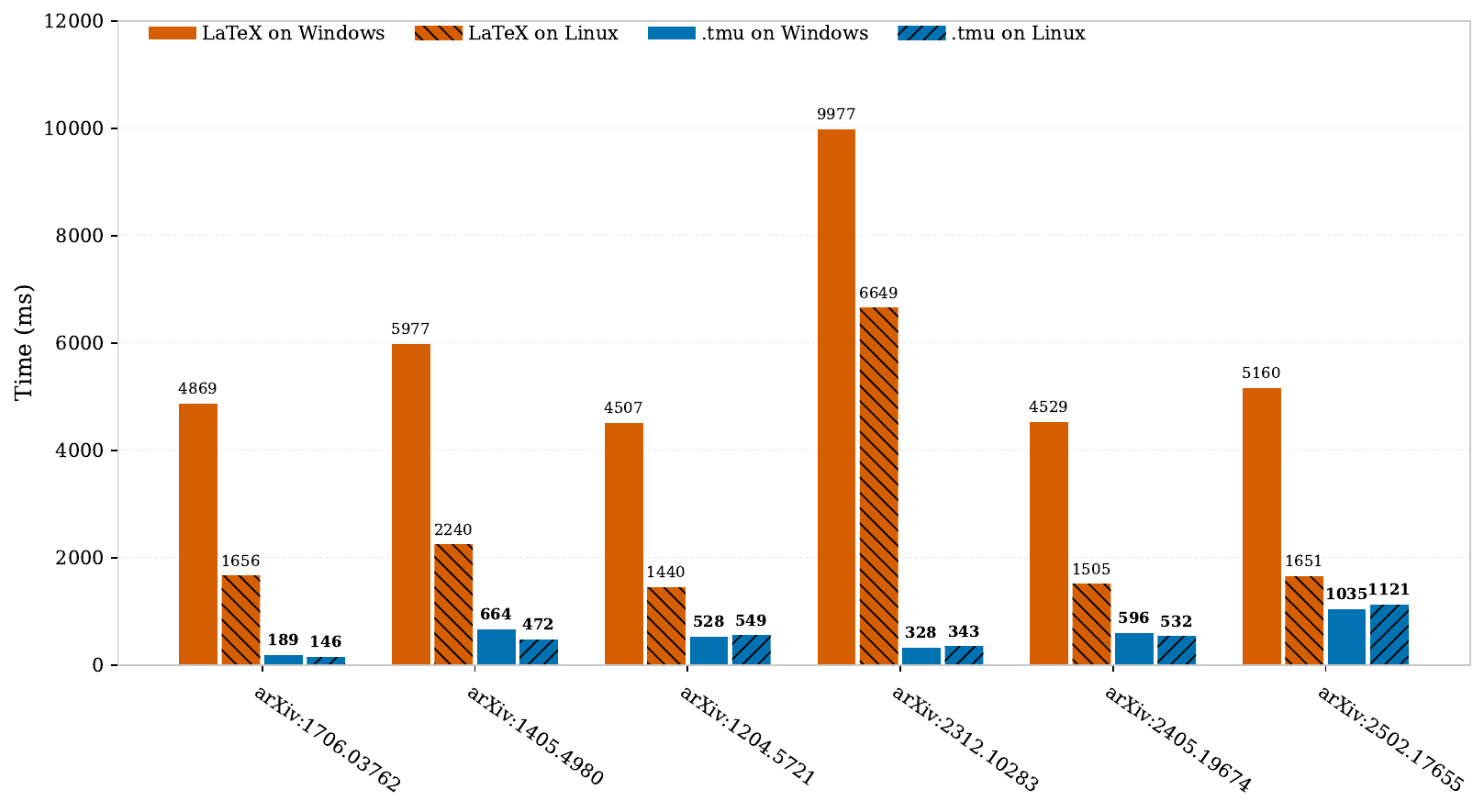}
    \caption{Benchmark on incremental update. Each bar represents the average of three trials.}
    \label{fig:inc-update}
\end{figure}

We benchmark compilation and rendering on 6 arXiv documents. \texttt{.tmu} outperforms \texttt{.tex} on both full compilation and incremental updates; the latter is most decisive (Figure~\ref{fig:inc-update}), because \texttt{.tmu} updates local tree nodes without I/O overhead ($t_{\text{IO}}=0$, activated only when opening the document). These low feedback delays directly reduce agent time cost in real-time LLM writing. \emph{(Full-compilation results and the $t_{\text{IO}}$ cost model are in Appendix~\ref{app:bench}.)}

\subsection{Performance in LLM tasks}

\begin{figure*}[t]
    \centering
    \begin{subfigure}[t]{0.32\textwidth}
        \centering
        \includegraphics[width=\linewidth]{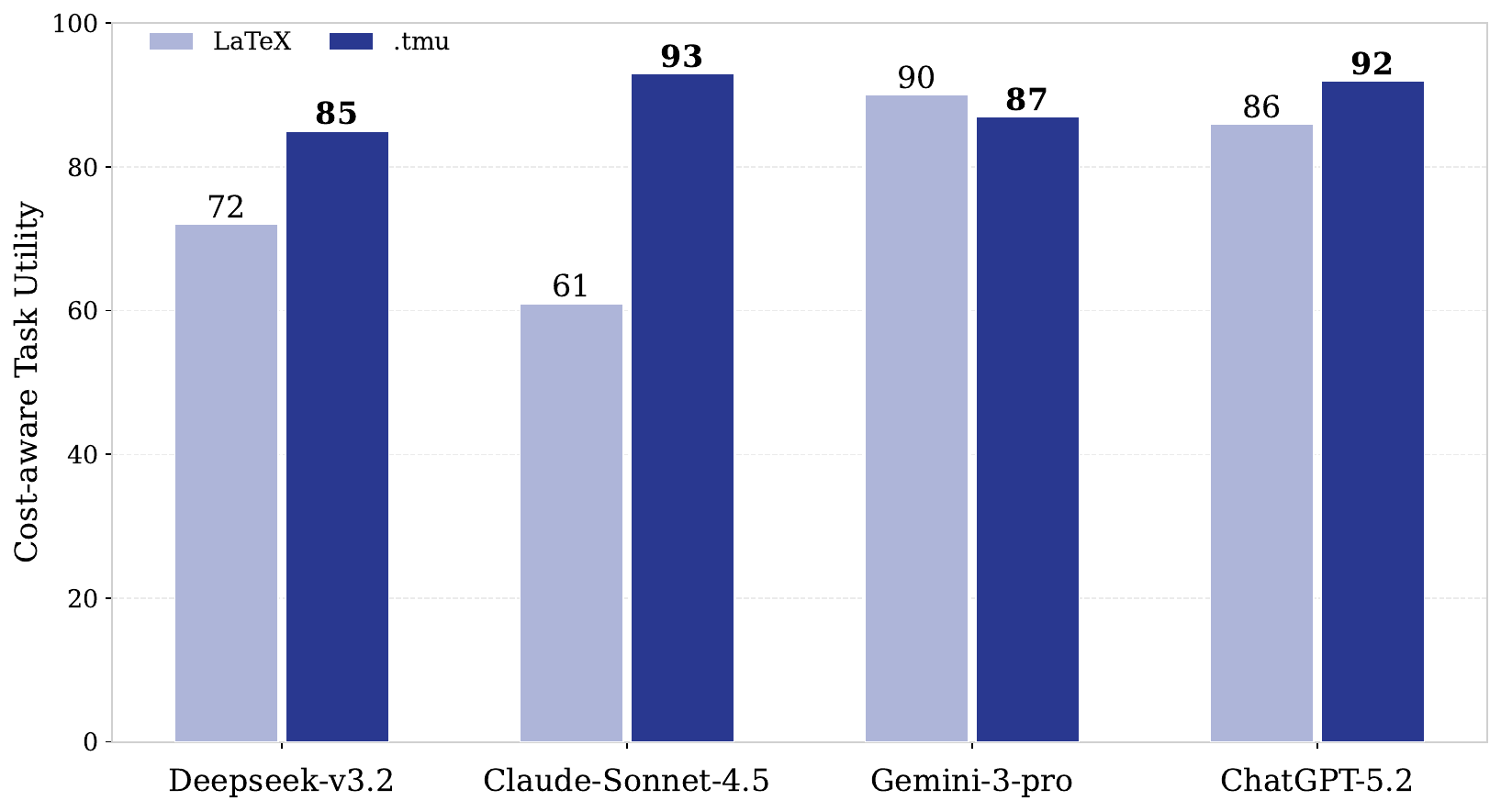}
        \caption{Structure locating}
        \label{fig:reading}
    \end{subfigure}\hfill
    \begin{subfigure}[t]{0.32\textwidth}
        \centering
        \includegraphics[width=\linewidth]{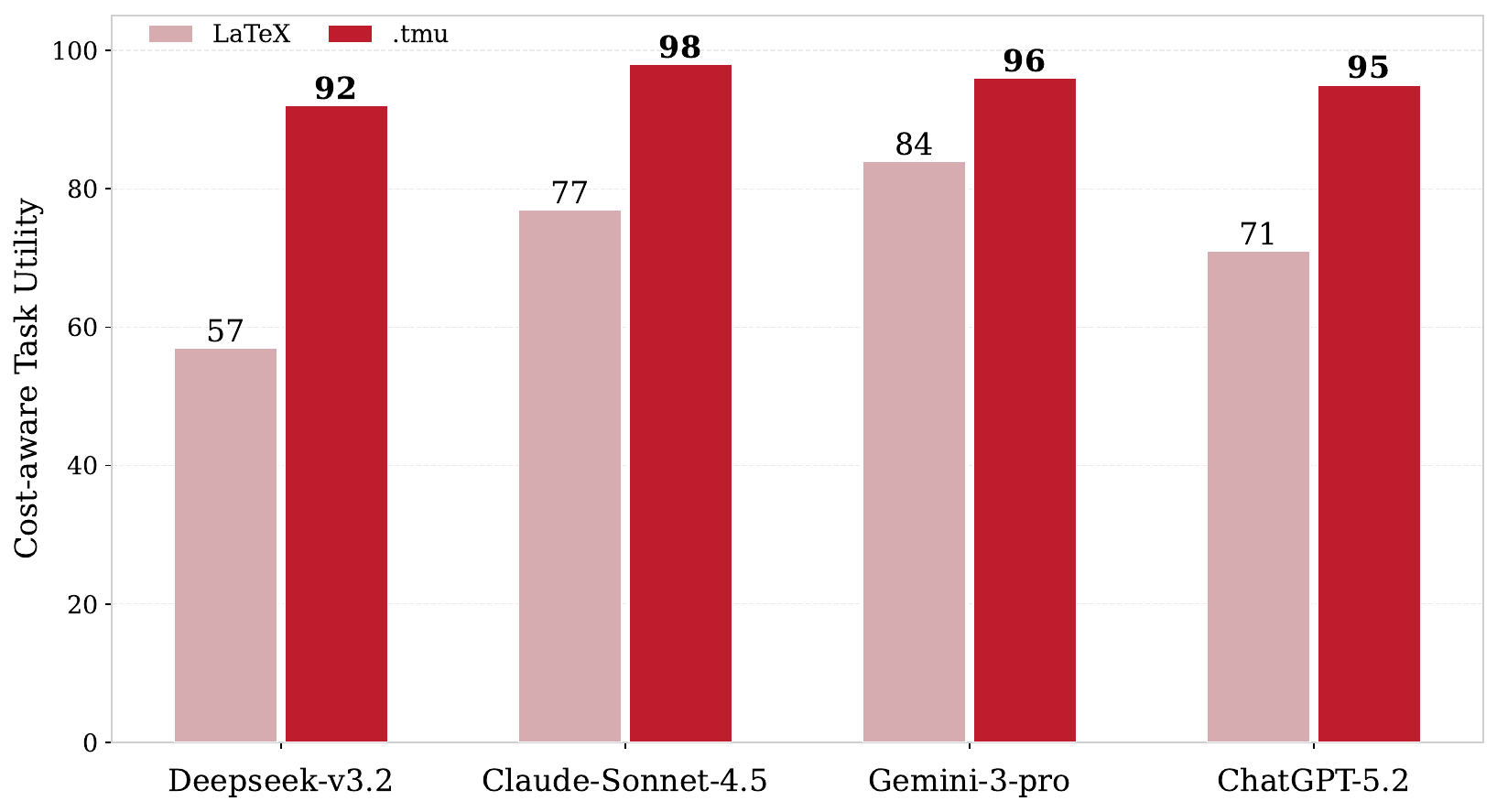}
        \caption{Style-aware merging}
        \label{fig:writing}
    \end{subfigure}\hfill
    \begin{subfigure}[t]{0.32\textwidth}
        \centering
        \includegraphics[width=\linewidth]{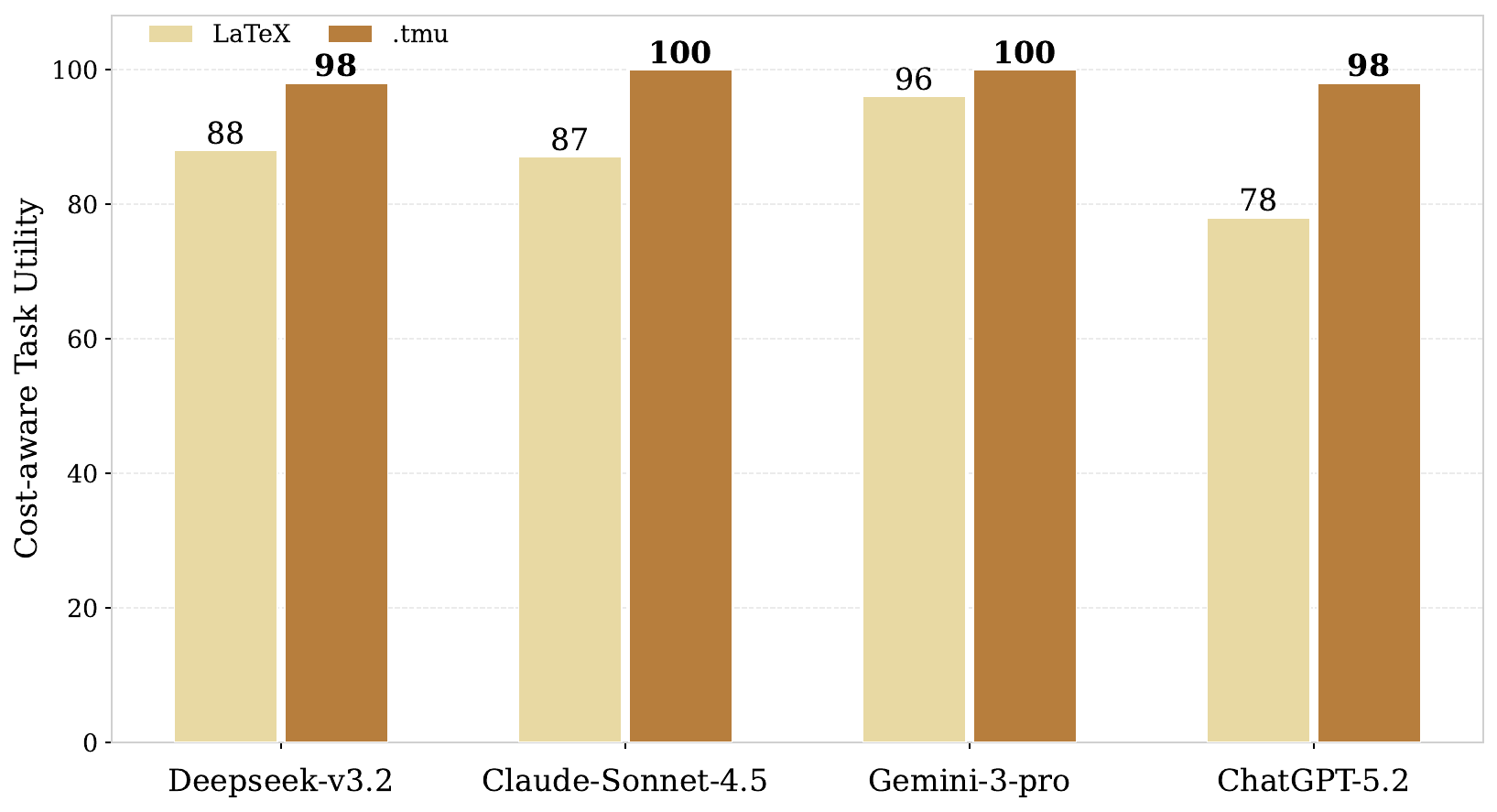}
        \caption{Debugging}
        \label{fig:debugging}
    \end{subfigure}
    \caption{LLM task performance for \texttt{.tmu} and \texttt{.tex}.}
    \label{fig:llm-tasks}
\end{figure*}

The standardized grammar and explicit tree-node tags in \texttt{.tmu} help models locate targets, merge contexts, and debug ill-formed structures more efficiently than \texttt{.tex}. Figure~\ref{fig:llm-tasks} summarizes the three LLM task results.

\subsubsection{Locating document structure}

To evaluate structural understanding, each LLM answers 20 document-structure questions (Appendix~\ref{app:20ques}). For each answer, we take
\begingroup\small
\[
    u_s = \max\left(0,\ \begin{cases}
        5 - \left\lfloor \dfrac{T}{1 \times 10^4} \right\rfloor & \text{right answer,} \\[6pt]
        0 & \text{wrong answer,}
    \end{cases} \right)
\]
\endgroup
where $\sum u_s \in [0, 100]$. The 10k-token scale separates confident structure location from longer reasoning/tool-use loops, which indicate lower efficiency and higher cost~\cite{ling_table2latex_2025}.

Despite LLMs' stronger pretraining exposure to \LaTeX{}, \texttt{.tmu} leads for most models (Figure~\ref{fig:reading}). Its reference metadata is updated and stored directly in the file (e.g., \path!<associate|sec:tree-struc-on-mogan|<tuple|5.1|13>>!), whereas \texttt{.tex} requires scanning and resolving the compiled numbering context.

Gemini is the only exception, scoring 90 on \texttt{.tex} versus 87 on \texttt{.tmu}. Thus, we claim gains in most, rather than all, settings and interpret this as a comparison of the complete native representations rather than tree syntax alone.

\subsubsection{Merging files with distinct doc-styles}
\label{sec:doc-style}

We define \emph{doc-style} as macro naming and command usage. Documents may use different theorem aliases, redefine the same macro differently, or give the same command incompatible meanings.

We ask LLMs to complete two assignments.

Assignment 1 generates two independently compilable files: \texttt{theorems.tex}, a theorem collection with many new macros and packages, and \texttt{proofs.tex}, a disordered proof collection with a distinct doc-style and conflicting packages. The files are connected by theorem and equation cross-references.

Assignment 2 merges the generated files into the doc-style of \texttt{theorems.tex}, while preserving compilability and placing each proof below the referenced theorem.

We obtain \texttt{theorems.tmu} and \texttt{proofs.tmu} from their \texttt{.tex} versions; rendered content is identical, as guaranteed by a \LaTeX{} importing engine.

Assignment 1 is 1 generation task; Assignment 2 is 4 merge tasks. For each task, we take
\begin{equation*}
\resizebox{\columnwidth}{!}{$
u_m =
\max\left(0,\
\begin{cases}
20 - 2E_{\mathrm{ref}} - \left\lfloor \dfrac{T}{10^4} \right\rfloor - E_{\mathrm{sty}}
    & \text{success on the first try,} \\[6pt]
10 - 2E_{\mathrm{ref}} - \left\lfloor \dfrac{T}{10^4} \right\rfloor - E_{\mathrm{sty}}
    & \text{success on the second try,} \\[6pt]
0
    & \text{fail within two tries,}
\end{cases}
\right)
$}
\end{equation*}
where $E_{\text{ref}}$ is the number of reference failures (i.e., "??" appears but compilation succeeds), $E_{\text{sty}}$ is the number of cases where the merged file has the doc-style of \texttt{proofs.tex} (we required LLMs to write the merged file in the doc-style of \texttt{theorems.tex}, so other macro aliases or redefined macros are not accepted), $\sum u_m \in [0, 100]$.

As shown in Figure~\ref{fig:writing}, \texttt{.tmu} scores higher for all LLMs. Its grammatical consistency removes many macro-conflict and style-unification choices that make complex \LaTeX{} generation brittle~\cite{kale2025texpert}; merging structured contexts becomes closer to copying compatible subtrees.
%

\subsubsection{Debugging ill-formed documents using error messages}

Debugging ill-formed documents is a common LLM co-writing use case. We construct 20 samples from the source document (Table~\ref{tab:ill-dist}), feed error messages to LLMs, and ask them to repair the document.

\begin{table}[t]
    \centering
    \small
    \setlength{\tabcolsep}{4pt}
    \caption{Distribution of error types.}
    \label{tab:ill-dist}
    \begin{tabular}{lc}
        \toprule
        \textbf{Error Type} & \textbf{Count} \\
        \midrule
        Unclosed bracket & 4 \\
        Unclosed environment & 5 \\
        Wrong command usage & 4 \\
        Undefined cross-reference & 3 \\
        Conflicting packages & 2 \\
        Self-recursive macros & 2 \\
        \bottomrule
    \end{tabular}
\end{table}


For each error, we take
\begingroup\small
\[
    u_d = \max\left(0,\ \begin{cases}
        5 - \left\lfloor \dfrac{T}{1 \times 10^4} \right\rfloor & \text{right answer,} \\[6pt]
        0 & \text{wrong answer,}
    \end{cases} \right)
\]
\endgroup
where $\sum u_d \in [0, 100]$.

 We treat this as an end-to-end evaluation under each format's native validation and diagnostic workflow, rather than a controlled comparison using identical plain-text error messages.

The \texttt{.tex} group requires substantially more reasoning on most samples, while the \texttt{.tmu} group solves all samples and only two exceed 10k tokens (Figure~\ref{fig:debugging}). A defect in \texttt{.tmu} affects only the closest tree-tag ancestor (Section~\ref{sec:tree-struc-on-mogan}), yielding localized messages instead of long logs detached from root causes.

Well-formed \texttt{.tmu} also precludes classes such as unclosed environments or self-recursive macros; when a malformed tree exists, a structured editor can surface the affected node directly while the rest of the document remains renderable.

\subsection{Efficiency in fine-tuning}
\label{sec:eff-in-sft}

Because \texttt{.tmu} is tree-structured while \texttt{.tex} follows a linear macro flow, next-token prediction should be less ambiguous in \texttt{.tmu}.

We fine-tune Qwen2.5-7B-Instruct~\cite{qwen2025qwen25} with LoRA on 1000 random formulas, converting each \LaTeX{} formula to an equivalent rendered \texttt{.tmu} S-expression. The formulas cover common mathematical constructs; each is split into a prefix prompt and completion target.

\begin{figure}[htbp]
    \centering
    \includegraphics[width=\columnwidth]{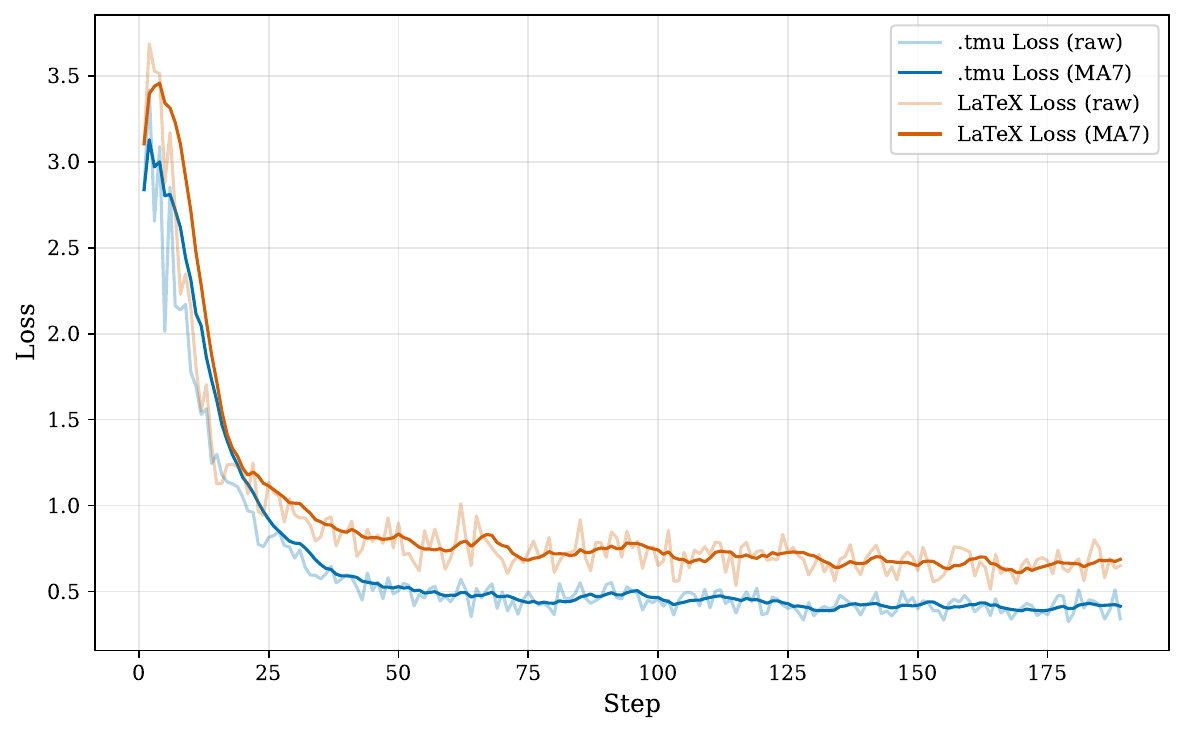}
    \caption{Experiment of LoRA based on Qwen2.5-7B-Instruct.}
    \label{fig:fine-tune}
\end{figure}

Following \citet{dong2025machinelearninglm}, both runs use identical LoRA settings and differ only in training format. Figure~\ref{fig:fine-tune} shows that \texttt{.tmu} converges to around 0.4 loss, while \texttt{.tex} converges to around 0.7. This supports the low-entropy claim: equivalent \LaTeX{} strings such as \verb|\frac{a}{b}| and \verb|{a \over b}|, or \verb|x^{2}| and \verb|x^2|, create extra prediction uncertainty absent from canonicalized \texttt{.tmu} trees.

This advantage matters especially for large documents with multiple doc-styles, where \LaTeX{} models must infer commands from preamble-local macro definitions~\cite{lin2024accurate}.

Moreover, discussions of the \texttt{.tmu} format versus Markdown are attached in Limitations.

\section*{Limitations}

\paragraph{Pretraining exposure and evaluation scope.} The LLMs we evaluate were pretrained on very large \LaTeX{} corpora and have seen essentially no \texttt{.tmu} during pretraining, placing \texttt{.tmu} at a substantial model-familiarity disadvantage. Evaluating models with better-matched exposure is future work. This scopes our evaluation, not our thesis: the architectural frictions {\LaTeX} poses for LLMs (one-off batch compilation, weak structural semantics, error messages detached from their root cause; Appendices~\ref{app:arch}--\ref{app:con}) are properties of the format, not artifacts of model familiarity, and remain our core motivation.

\paragraph{Evaluation scale and statistics.} The LLM-task evaluation is focused rather than broad: ${\sim}20$ items each for structure-locating and debugging, only a handful of merging tasks, drawn largely from a single source document (arXiv:2502.17655). We report point estimates without significance testing or variance, and do not yet claim domain generality. A larger, multi-domain benchmark with repeated trials and significance testing is future work.

\paragraph{Scope of the fine-tuning evidence.} The fine-tuning result uses a single LoRA run on one model family (Qwen2.5-7B-Instruct), a single seed, and formula-level rather than full-document data. Lower training loss is consistent with but does not by itself establish better downstream generation; validation loss, exact-match and rendered-equivalence metrics, and multi-seed runs across model sizes are future work.

\paragraph{Utility metric.} Our utility scores ($u_s$, $u_m$, $u_d$) combine task correctness with token cost under a hand-chosen 10k-token scale; absolute scores are not comparable across task types, and the metric entangles accuracy with token efficiency. Reporting these two axes separately is future work.

\paragraph{Baselines and the tool-vs-representation confound.} Our baselines are limited to {\LaTeX}; comparison against Typst, Markdown-with-math, and MathML/XML and other structured editors is future work. We note Markdown specifically because it reuses {\LaTeX} math syntax and is the natural lightweight baseline, but a fair comparison is non-trivial given the gap in intended use, so we do not treat it as central. A further confound is that some gains may stem from the editor's tooling (incremental update, structured editing, the {\LaTeX} import engine) rather than the representation alone; disentangling these requires controlled ablations we have not yet run.

\paragraph{On low entropy.} We currently support the low-entropy claim with an operational proxy (reduced source-encoding multiplicity for a fixed rendered output, e.g.\ \texttt{x\^{}2} vs.\ \texttt{x\^{}\{2\}}) together with indirect evidence from fine-tuning convergence. A full information-theoretic treatment is future work.

\paragraph{Future work.} A concrete agenda: (i) a fair controlled comparison with an ablation separating the contribution of tree structure from that of pre-expanded reference numbers, isolating the structure-locating input asymmetry (in \texttt{.tmu}, numbering is stored inline after incremental update, whereas {\LaTeX} requires a full scan); (ii) a larger multi-domain evaluation with repeated trials, significance testing, and additional baselines (Typst, Markdown-with-math, MathML); (iii) fine-tuning with validation loss, exact-match, and multi-seed runs across model sizes, plus a Markdown comparison and a larger, standardized end-to-end LLM-repair benchmark.



\bibliography{references}

\clearpage
\appendix
\section{A Brief History of \TeX}
\label{app:intro}

Numerous derivatives of \TeX{} have emerged in the past decades, with \LaTeX{} being the most prominent. Built as a macro language on top of \TeX{}, \LaTeX{} significantly simplifies its usage, allowing users to leverage \TeX{}'s powerful typesetting capabilities without needing a deep understanding of intricate commands. By defining commands and templates that align with standard typesetting practices, \LaTeX{} has made the production of scientific literature and books far more efficient and accessible, eventually becoming the de facto standard for scientific document preparation.

In the academic field, the \TeX{} system and \LaTeX{} in particular have become the standard of the scientific community thanks to their exceptional mathematical typesetting capabilities. The core design of \TeX{} originates from the pioneering work of \citet{knuth1984texbook}. The American Mathematical Society (AMS) strongly encourages mathematicians to submit manuscripts using \TeX{}, and widespread adoption by world-class publishers such as Wesley and IEEE has made it a staple for books and journals. Consequently, \TeX{} occupies a pivotal position in the production of academic papers and monographs, serving as a vital tool for scholarly communication and knowledge dissemination.

However, \TeX{}'s original design and its subsequent development trajectory have resulted in numerous legacy issues. This article primarily examines the underlying architecture of \TeX{} and explains why certain design choices have led to significant problems.

On another note, it was long believed that ``What You See Is What You Get'' (WYSIWYG) was incompatible with structured editing. The emergence of \TeX{}macs, however, proved this assumption fundamentally incorrect. Professor Joris from \'Ecole Polytechnique wrote a critique on this subject (see \citealp{van_der_hoeven_gnu_2001, liiistem2025}). The design philosophy of \LaTeX{} has inspired a series of similar editing software; beyond the aforementioned \TeX{}macs, these include LyX and the recently popular Typst.

Notably, the formula rendering in Typst and \TeX{}macs is completely independent of the \TeX{} system, whereas LyX serves as a front-end for \TeX{}. While this article contains significant criticism of the \TeX{} system, we wish to clarify our stance: we are by no means denying \TeX{}'s historical status, nor do we suggest it was outdated for its time. We simply argue that today---especially in an era of rapidly advancing AI tools---\TeX{}'s underlying design presents many issues.

\section{An Introduction to \TeX{} Compilation Principles}
\label{app:pr}

\begin{lstlisting}[caption={Minimal example of \LaTeX{} code},label={lst:minimal}]
\documentclass{article}
\usepackage{siunitx}
\begin{document}
Hello, World! from \LaTeX.
The speed of light is \SI{299792458}{\meter\per\second}.
\end{document}
\end{lstlisting}

Listing~\ref{lst:minimal} shows a minimal example of \LaTeX{} code. The source code begins with the \verb|\documentclass| command. Lamport's \LaTeX{} framework simplifies the use of \TeX{}~\cite{lamport1994document} by defining the document class used for the file. Immediately following this, we can use \verb|\usepackage| to import packages. We then use \verb|\begin{document}| and \verb|\end{document}| to mark the start and end of the body content, placing the text between them.

The area between \verb|\documentclass| and \verb|\begin{document}| is called the \textit{preamble}. In addition to importing packages via \verb|\usepackage|, \textit{preamble} is used for global macro definitions and document configuration, which can also be left empty.

Finally, we can compile the code. Running \verb|pdflatex example.tex| generates the \verb|example.pdf| file. Simultaneously, several auxiliary files are generated in the same directory. For instance, the \verb|.aux| file saves intermediate information such as cross-references, tables of contents, and numbering; consequently, documents often require multiple compilations to yield correct results. The \verb|.log| file records the complete compilation process and is the primary resource for locating warnings and errors. The \verb|.dvi| file stands for \textit{DeVice Independent file}, an intermediate format unrelated to the specific output device.

Beyond the \verb|.tex| source files we write, every package and document class consists of files with specific extensions. Table~\ref{tab:latex-ext} lists the files that frequently appear in \LaTeX{} templates:

\begin{table}[htbp]
    \centering
    \small
    \setlength{\tabcolsep}{3pt}
    \caption{\LaTeX{} file extensions and their descriptions}
    \label{tab:latex-ext}
    \begin{tabular}{p{1cm} p{2.2cm} p{3.8cm}}
        \toprule
        \textbf{Ext.} & \textbf{File Type}   & \textbf{Description}                                                       \\
        \midrule
        \texttt{.sty}           & Package file         & The package name corresponds to the filename without the extension.        \\
        \texttt{.cls}           & Document class file  & The document class name corresponds to the filename without the extension. \\
        \texttt{.bib}           & BibTeX database file & A database file storing bibliographic information.                         \\
        \texttt{.bst}           & BibTeX style file    & A template file defining the formatting style for bibliographies.          \\
        \bottomrule
    \end{tabular}
\end{table}

\LaTeX{} generates numerous auxiliary files and logs during the compilation process. Features such as cross-references, bibliographies, tables of contents, and indices require an initial compilation to generate auxiliary files, followed by a subsequent compilation to read these files to produce the correct result. Therefore, complex \LaTeX{} source code often requires multiple compilation passes as summarized in Table~\ref{tab:latex-aux}:

\begin{table}[htbp]
    \centering
    \small
    \setlength{\tabcolsep}{3pt}
    \caption{\LaTeX{} auxiliary file extensions, tools, and descriptions}
    \label{tab:latex-aux}
    \begin{tabular}{p{1cm}p{2.2cm}p{3.6cm}}
        \toprule
        \textbf{Ext.} & \textbf{Related Tool / Source} & \textbf{Description}                                                \\
        \midrule
        \texttt{.log}      & Typesetting Engine             & Records the compilation process; used for error debugging.          \\
        \texttt{.aux}      & \texttt{latex}                 & Main auxiliary file; records cross-references, TOC, citations, etc. \\
        \texttt{.toc}      & \texttt{latex}                 & Records content for the Table of Contents.                          \\
        \texttt{.lof}      & \texttt{latex}                 & Records content for the List of Figures.                            \\
        \texttt{.lot}      & \texttt{latex}                 & Records content for the List of Tables.                             \\
        \texttt{.bbl}      & \texttt{bibtex}                & The generated bibliography record file.                             \\
        \texttt{.blg}      & \texttt{bibtex}                & Log file for the BibTeX compilation process.                        \\
        \texttt{.idx}      & \texttt{latex}                 & Raw index record file to be processed by \texttt{makeindex}.        \\
        \texttt{.ind}      & \texttt{makeindex}             & Formatted index file generated after processing \texttt{.idx}.      \\
        \texttt{.ilg}      & \texttt{makeindex}             & Log file for \texttt{makeindex}.                                    \\
        \texttt{.out}      & \texttt{hyperref} package      & Record file for generating PDF bookmarks.                           \\
        \bottomrule
    \end{tabular}
\end{table}

To intuitively understand why \LaTeX{} documents often require multiple compilations, Listing~\ref{lst:crossref} shows a minimal example containing cross-references and a table of contents:

\begin{lstlisting}[caption={Cross-reference example},label={lst:crossref}]
\documentclass{article}
\usepackage{hyperref}

\begin{document}
\tableofcontents

\section{Introduction}\label{sec:intro}
See Section~\ref{sec:intro}.
\end{document}
\end{lstlisting}

When compiling this code with \verb|pdflatex| for the first time, \LaTeX{} cannot yet determine section numbers or the content of the table of contents (TOC). The compiler records this information into auxiliary files during the typesetting process: section titles are written to the \verb|.toc| file, and numbering information for \verb|\label{sec:intro}| is written to the \verb|.aux| file. Meanwhile, the \verb|\ref{sec:intro}| in the body text is temporarily output as a placeholder. Consequently, the PDF generated from the first pass has an empty TOC, and cross-references appear as "??".

During the second pass, \LaTeX{} reads the \verb|.aux| and \verb|.toc| files generated previously, obtaining complete numbering and directory information to typeset them correctly into the document. Thus, the cross-references appear correctly, and the TOC is populated. This process demonstrates that the \LaTeX{} typesetting process is essentially an iterative workflow that relies on intermediate results to converge. Any feature involving global information or forward/backward dependencies almost inevitably requires multiple compilation passes to produce the final correct output.

In practice, the typesetting result is determined by the underlying engine. Common engines include:

\begin{itemize}
    \item \textbf{pdfLaTeX}: The most traditional and compatible engine. It generates PDFs directly but has limited support for Unicode and system fonts, relying mostly on \TeX{}'s own font system.
    \item \textbf{XeLaTeX}: Renowned for excellent Unicode support and the ability to call operating system fonts directly. It is highly suitable for multi-language typesetting (such as CJK), though its compilation speed and package compatibility are often more complex than pdfLaTeX.
    \item \textbf{LuaLaTeX}: Builds upon XeLaTeX by introducing Lua scripting as a programmable extension layer. This mechanism allows for dynamic customization of typesetting logic, offering the most power but also the highest complexity, with a strong dependency on the quality of templates and packages.
\end{itemize}

Differences in font handling, package support, and compilation behavior among these engines mean that the same \LaTeX{} source code frequently yields different results---or fails to run entirely---in different environments. The coexistence of these diverse engines is one of the root causes of the long-standing complexity within the \LaTeX{} ecosystem.

\section{Fundamental Defects in \TeX{}'s Compilation Design}
\label{app:arch}
\label{sec:fundamental}

The batch processing compilation model adopted by the \TeX{} system was rational within the computing environment of its inception in the 1970s and 80s. However, in the context of contemporary academic writing, which emphasizes interactivity, instant feedback, and multi-platform publishing, this model exposes profound systemic tension. Its core issues lie in its \textit{one-off processing flow}, \textit{weak semantic representation}, and \textit{delayed error feedback mechanisms}. These problems not only directly impair user experience but fundamentally constrain the evolution of the tool ecosystem.

\subsection{Batch Model Limitations: Unidirectionality, Weak Semantics, and Delayed Feedback}

\TeX{}'s compilation architecture is rooted in the batch processing paradigm. While capable of efficiently handling static documents, it struggles to adapt to the dynamic, iterative needs of modern writing and LLM training~\cite{paster2023openwebmath}.

\subsubsection{Unidirectional and One-off Processing Flow}

The working mechanism of \TeX{} follows a rigid sequence: Linear Input Reading $\rightarrow$ Macro Expansion $\rightarrow$ Typesetting Calculation $\rightarrow$ Output Generation.

This flow possesses strong unidirectionality and irreversibility. The system does not maintain an intermediate representation amenable to incremental re-computation; instead, it treats input as a continuous stream of instructions. This design leads to:

\textbf{Delayed Manifestation of Global State:} The overall typesetting state is only evaluable after the full compilation cycle concludes. \TeX{}'s typesetting decisions (pagination, float positioning, citation numbering, etc.) rely on global information, which is often only determined after the document has been fully read and executed. Consequently, users cannot verify whether the document is "typeset correctly" before compilation is complete. Listing~\ref{lst:xref-before-def} shows a typical example:

\begin{lstlisting}[caption={Cross-reference before definition},label={lst:xref-before-def}]
\documentclass{ctexart}
\begin{document}

As shown in Equation~\ref{eq:test}.

\newpage

\begin{equation}
  E = mc^2
\label{eq:test}
\end{equation}

\end{document}
\end{lstlisting}

When this code is compiled with \texttt{xelatex} for the first time, \texttt{\textbackslash ref\{eq:test\}} displays as \texttt{??}. This discrepancy occurs because the equation's numbering information remains ungenerated; \TeX{} cannot know the correct reference number during the current compilation pass. Only after the first pass concludes and the numbering information is written to the \texttt{.aux} file can \TeX{} read this information during the second compilation, enabling \texttt{\textbackslash ref\{eq:test\}} to display the correct number.

This requirement for multiple passes demonstrates that \TeX{}'s typesetting decisions are global and latent: the typesetting engine cannot accurately infer cross-references or numbering within the document before the first compilation finishes. Multiple passes are strictly required to obtain the final result.

\textbf{Inability to Implement Incremental Updates:} Local modifications cannot trigger efficient, localized re-computation; instead, the entire compilation process requires repetition. Even if only a single local element in the document is modified, \TeX{} must re-execute the entire input stream from the beginning, re-expanding macros, re-paginating, and re-calculating all layouts. Listing~\ref{lst:incremental} illustrates this issue:

\begin{lstlisting}[caption={Incremental update example},label={lst:incremental}]
\documentclass{ctexart}
\usepackage{lipsum}
\begin{document}

\section{Section 1}
Some text, some text, some text.
\lipsum[2-8]
\section{Section 2}
More text, more text, more text.

\end{document}
\end{lstlisting}

Suppose we delete line 7, \texttt{\textbackslash lipsum[2-8]}, which originally generated multiple paragraphs occupying significant vertical space. Keeping the rest of the code unchanged, semantically, this is merely a local modification to the internal content of the first section; it does not touch the structure of the second section or anything following it.

However, during actual compilation with \texttt{xelatex}, this modification triggers a cascading reaction: the vertical space previously occupied by \texttt{\textbackslash lipsum[2-8]} vanishes, drastically reducing the height of the first section. This alters the pagination results for the entire document. The title of the second section, which might have been on the next page, moves up to the previous page; consequently, page numbers, headers, footers, and the positions of any floating bodies must be recalculated. Despite the modification occurring in just a single line at the beginning of the document, \TeX{} is compelled to re-read the input stream from the start, re-expand all macros, and re-execute the complete pagination algorithm. It is unable to perform a local reflow solely for "Section 1" or the "affected pages."

This example clearly demonstrates that within \TeX{}'s execution model, there is no stable, reusable intermediate typesetting state. Any seemingly local change can alter all subsequent typesetting decisions. Therefore, the system has no choice but to perform a full re-compilation, making the efficient incremental updates expected of modern editors impossible.

\textbf{Lack of Visualization Upon Interruption:} Once compilation is interrupted by an error, users cannot obtain reliable partial results for preview. \TeX{} typically aborts the output process immediately, preventing users from seeing "what has been typeset so far." Listing~\ref{lst:missing-arg} demonstrates this scenario:

\begin{lstlisting}[caption={Missing argument error},label={lst:missing-arg}]
\documentclass{ctexart}
\newcommand{\mycmd}[1]{\textbf{#1}}
\begin{document}
This page contains perfectly correct content.

\mycmd % Forgot to provide argument

The content following this theoretically does not
depend on this command.
\end{document}
\end{lstlisting}

When this document is compiled with \texttt{xelatex}, the process fails to complete, generating only \texttt{.aux} and \texttt{.log} files. The error prompt is \texttt{Runaway argument?}. This occurs because \TeX{} detects an incomplete or missing argument during macro processing (in this case, \texttt{\textbackslash mycmd} lacks a necessary argument), causing macro expansion to exceed its expected scope. Consequently, subsequent content cannot be typeset, and the entire PDF output fails. Users are forced to analyze the log to locate the error, unable to verify the pages that were already successfully typeset.

This characteristic stands in sharp contrast to the incremental update and continuous feedback mechanisms prevalent in modern editors, significantly constraining user iteration efficiency.

\subsubsection{Tight Coupling Between Compilation and Semantic Phases}

In the \TeX{} system, syntax parsing, macro expansion, semantic interpretation, and typesetting decisions do not form a clear stratified structure; instead, they are deeply coupled within a single execution path. Macro expansion assumes the role of "semantic modeling" while simultaneously triggering specific typesetting behaviors, thereby conflating abstract structure with layout details. While this design offered high flexibility in the early days, it has exposed serious deficiencies in the context of modern document engineering.

First, macro commands often bear the dual responsibility of structural semantics and layout control. Taking \texttt{\textbackslash section} as an example, this command should logically represent only the structural semantic of "section hierarchy." However, its implementation dictates the font size, spacing, numbering method, and whether a table of contents entry is generated as shown in Listing~\ref{lst:section-dual}:

\begin{lstlisting}[caption={Section command with dual responsibility},label={lst:section-dual}]
\section{Introduction}
\end{lstlisting}

This single line of code not only declares a new structural node but immediately triggers a series of typesetting side effects, including incrementing counters, formatting the title, writing to the TOC (\texttt{.toc}), and generating bookmarks (if \texttt{hyperref} is loaded). Because these actions occur simultaneously during the macro expansion phase, downstream tools cannot distinguish whether a command is a "structure declaration" or a "concrete typesetting instruction," making it difficult to understand the document structure without executing the code.

Second, the macro definition mechanism itself cannot distinguish between abstract semantics and concrete implementation. This results in the same syntactic form potentially representing either high-level semantics or merely a typesetting shortcut. For example: \texttt{\textbackslash def\textbackslash theorem\#1\{\textbackslash textbf\{Theorem.\} \#1\}}. Superficially, \texttt{\textbackslash theorem} appears to be a semantic structure (a theorem). In reality, it is a simple wrapper for bold text, lacking numbering, cross-referencing, or hierarchical information. In contrast: \texttt{\textbackslash newtheorem\{theorem\}\{Theorem\}} introduces true structural semantics, including automatic numbering, scoping, and citability. However, from the perspective of the \TeX{} engine, there is no essential difference between these two definitions during the macro expansion phase---both are simply "executable text substitution rules." This semantic indistinguishability prevents static analysis tools from determining whether a specific macro represents document structure or merely affects visual appearance.

Furthermore, the conditional expansion mechanism directly influences the final structure of the document during the parsing phase, rather than merely acting as a late-stage typesetting choice. Listing~\ref{lst:conditional-structure} shows an example:

\begin{lstlisting}[caption={Conditional structure},label={lst:conditional-structure}]
\ifdefined\includeappendix
  \section{Appendix}
\fi
\end{lstlisting}

Whether the structural node "Appendix" exists depends entirely on the truth value of a condition at the moment of macro expansion. In other words, the document's logical structure is not a stable, enumerable abstract tree, but a result dynamically generated during execution. Any tool attempting to analyze the document structure without running \TeX{} must completely simulate macro expansion and conditional logic, which is virtually unfeasible in practice.

The aforementioned high degree of coupling leads to multifaceted negative impacts. First, static analysis tools find it difficult to intervene. Due to the lack of clear semantic boundaries, tools cannot reliably construct an Abstract Syntax Tree (AST) or structural model, making advanced features like code refactoring, semantic checking, and consistency verification arduous to implement.

Since \LaTeX{} compilation results are strongly dependent on instruction order, \LaTeX{} introduced engineering improvements to allow users or package authors to insert and execute custom code at specific execution timing points. These mechanisms are not based on an explicit semantic event model, but rather attach to \TeX{}'s sequential execution flow, coordinating behavior between packages and document structure by "intercepting execution timing." This enables functions such as lazy initialization, global parameter patching, auxiliary file processing, and layout adjustment. In the \LaTeX{} core system, the commonly used hook commands primarily include the four types listed in Table~\ref{tab:latex-hooks}:

\begin{table}[htbp]
  \centering\textit{}
  \small
  \setlength{\tabcolsep}{3pt}
  \caption{Common \LaTeX{} Hook Directives and Descriptions}
  \label{tab:latex-hooks}
  \begin{tabular}{p{2.8cm}p{4.0cm}}
    \toprule
    \textbf{Directive}                      & \textbf{Description}                                                                                   \\
    \midrule
    \texttt{\textbackslash AtBeginDocument} & Executes code at the beginning of the document body (after \texttt{\textbackslash begin\{document\}}). \\
    \texttt{\textbackslash AtEndDocument}   & Executes code before the document ends (before \texttt{\textbackslash end\{document\}}).               \\
    \texttt{\textbackslash AtEndOfPackage}  & Executes code immediately after the current package finishes loading.                                  \\
    \texttt{\textbackslash AtEndOfClass}    & Executes code immediately after the current document class finishes loading.                           \\
    \bottomrule
  \end{tabular}
\end{table}

These directives provide a relatively non-invasive method of collaboration between packages and user documents, avoiding the need to forcibly control loading order or directly modify user source code to achieve functional extension.

\subsubsection{Lagged Manifestation and Ambiguous Localization of Errors}

\TeX{}'s error detection mechanism is fundamentally an \textbf{"execution-driven, post hoc"} paradigm. This is a direct consequence of its compilation model, where linear macro expansion is inextricably interwoven with immediate typesetting. Since the system does not maintain a structured intermediate state capable of real-time verification during execution, errors manifest passively only when execution becomes impossible, exhibiting a set of systemic characteristics.

First, error types are highly conflated at the diagnostic level. Structural defects at the semantic level (e.g., mismatched environments, abnormal parameter expansion) are often reported indistinguishably from failures at the typesetting level (e.g., box overflows, mode-switching errors). Consequently, error messages fail to reflect the relevant level of abstraction, significantly increasing debugging difficulty.

Second, the reported location of an error is rarely its actual point of origin. The line numbers and context provided by \TeX{} typically correspond to the point where the system finally "crashed" or stalled, rather than the source where the issue was introduced. Furthermore, the point of interruption and the origin of the error may be separated by a significant distance: an early, local error might only be exposed in a completely different form after multiple cycles of macro expansion and typesetting decisions. The result is a globalization of local issues, where a subtle defect in a single environment or command is sufficient to cause an irreversible interruption of the entire compilation process.

It must be emphasized that this is not an oversight in specific implementation details, but a characteristic rooted in the fundamental design of \TeX{}: the system lacks a structured state model independent of the execution process that enables continuous consistency checking. In other words, \TeX{} possesses no "prior knowledge" regarding the document's validity; it reports failure only when execution becomes untenable. Listing~\ref{lst:nested-section} shows a confusing error case:

\begin{lstlisting}[caption={Nested section commands},label={lst:nested-section}]
\documentclass{ctexart}
\begin{document}
\section{aaa \subsection{bbb}}
\end{document}
\end{lstlisting}

From the perspective of document structure, this usage of \texttt{\textbackslash section} is clearly illegal: a subsection is nested within the parameter of a section title. However, \LaTeX{} does not report this as a "structural nesting error." Instead, the system enters an inconsistent state during the macro expansion and typesetting execution, eventually throwing a confusing error message such as "LaTeX Error: Not allowed in LR mode." This message neither identifies the violating structural relationship nor points to the location where the issue actually occurred.

The following examples highlight a typical characteristic of \TeX{}'s error detection mechanism: a significant offset often exists between the true origin of the error and the reported location. The unclosed-brace example promoted to the main text (Listing~\ref{lst:unclosed-frac}) illustrates this issue: the second argument of \texttt{\textbackslash frac} lacks a closing brace. Semantically, this is a distinct local structural error. Yet, \TeX{} does not immediately report an error upon reading this line; instead, it continues to treat subsequent input as part of the argument, maintaining math mode and attempting to complete macro expansion. It is not until \texttt{\textbackslash end\{equation\}} is reached that the system detects the failure of the grouping and mode states to converge properly, triggering a fatal "Runaway argument?" error. By this point, the reported location is far removed from the actual occurrence, forcing users to manually backtrack through the context to verify closed groups and locate the root cause.

A similar latency phenomenon occurs with mismatched delimiters, such as using \texttt{\textbackslash left(} while omitting the corresponding \texttt{\textbackslash right)}. In this scenario, \TeX{} persists in waiting for a command to close the extensible delimiter. Superficially, the formula continues to be accepted and parsed, but its internal state is already inconsistent. Ultimately, the error is triggered at \texttt{\textbackslash end\{equation\}}, potentially interrupting compilation with a message like "You can't use \texttt{\textbackslash eqno} in math mode." It is crucial to emphasize that such errors do not directly point to the logical cause ("missing \texttt{\textbackslash right}"); rather, they reflect the inability to finalize the typesetting process at the closing stage. This transforms an issue of local structural incompleteness into a failure at the mode or typesetting level.

Both scenarios illustrate a fundamental constraint: \TeX{} does not maintain a structural state model capable of immediate verification. Instead, it relies on the passive exposure of issues when the execution process fails. This is one of the root causes of its "lagged manifestation and ambiguous localization" of errors.

\begin{remark}
  The \texttt{.tmu} format effectively mitigates this issue at the structural level. By representing content with implicit \texttt{around} tags and explicit structural wrapping, the format maintains a verifiable structural state during the input phase, preventing error propagation. See Section~\ref{sec:tree-struc-on-mogan} for details.
\end{remark}

\subsection{Compatibility Dilemma Beneath a Unified Syntax}

To address varying requirements (such as modern fonts and Unicode support), the \LaTeX{} ecosystem has evolved multiple parallel compilation engines, such as \texttt{pdfLaTeX}, \texttt{XeLaTeX} and \texttt{LuaLaTeX}. This differentiation, rather than being a strength, is a manifestation of system design fragmentation.

From a user's perspective, all three engines accept \texttt{.tex} input and adhere to the \LaTeX{} macro interface. However, at the system level, they exhibit fundamental differences in the key aspects summarized in Table~\ref{tab:latex-engines}:

\begin{table*}[htbp]
  \centering
  \caption{Comparison of \LaTeX{} Engines}
  \label{tab:latex-engines}
  \resizebox{0.99\textwidth}{!}{%
    \begin{tabular}{llll}
      \toprule
      \textbf{Dimension}       & \textbf{pdfLaTeX}                  & \textbf{XeLaTeX}             & \textbf{LuaLaTeX}  \\
      \midrule
      Native Encoding Support  & 8-bit (requires \texttt{inputenc}) & Native Unicode               & Native Unicode     \\
      Font System              & Type1 / limited OpenType           & System Fonts (OpenType)      & OpenType (via Lua) \\
      Extension Mechanism      & \TeX{} primitives                  & \TeX{} + Xe\TeX{} primitives & \TeX{} + Lua VM    \\
      External Programmability & Very Weak                          & Very Weak                    & Strong (Lua)       \\
      \bottomrule
    \end{tabular}%
  }
\end{table*}

These discrepancies imply that the same \texttt{.tex} source file does not correspond to the same semantic execution environment across different engines.

\textbf{Compatibility Burden:} A single document may require switching between engines to resolve specific issues, forcing users to understand the subtle differences and limitations of each engine.

Take the use of Emojis as an example. If a document requires the direct inclusion of Unicode Emojis (e.g., 🙂📊), \texttt{pdfLaTeX} is mechanically incapable of supporting such characters.\footnote{Some packages such as \texttt{hwemoji} attempt to work around this limitation by pre-compiling emojis into PDF resources and replacing them during compilation. However, this is not native support---i.e., direct rendering from text and font files---but rather a workaround. We still consider \texttt{pdfLaTeX} to be fundamentally incapable of supporting such characters.} While \texttt{XeLaTeX} possesses native Unicode support, in practice, it often degrades Emojis to monochrome glyphs or suffers from missing characters due to insufficient font coverage. \texttt{LuaLaTeX} offers the theoretically most complete support path, handling complex characters via OpenType fonts and the Lua layer. However, this 'enhanced capability' comes at a price: \texttt{LuaLaTeX}'s compilation speed is typically significantly slower than \texttt{XeLaTeX}, and it introduces an additional dependency on the Lua runtime, expanding both the source of errors and the scope of debugging. Under these constraints, a compromise workflow that actually exists in reality involves: compiling the main body with \texttt{XeLaTeX} for speed and layout stability; separately compiling pages or chapters containing Emojis using \texttt{LuaLaTeX}; and finally stitching the output results from different engines together at the PDF level.

A similar compatibility burden appears in the bibliography processing chain. While \LaTeX{} superficially provides a unified \texttt{.bib} data source format, during the actual compilation process, users must choose between different backends like \texttt{bibtex} and \texttt{biber}. These two differ fundamentally in character encoding, sorting rules, and data models as shown in Listing~\ref{lst:bibtex-biber}:

\begin{lstlisting}[caption={BibTeX vs BibLaTeX},label={lst:bibtex-biber}]
\bibliography{ref/refs}   % References compiled using BibTeX
% \printbibliography       % References compiled using BibLaTeX/Biber
\end{lstlisting}

Visually, these commands differ by only a single line, yet they correspond to two nearly incompatible toolchains. The former triggers the traditional \texttt{bibtex} workflow: the \texttt{.bib} file is parsed by \texttt{bibtex} to generate a \texttt{.bbl} file, employing a data model constrained by 8-bit encoding and a rigid \texttt{.bst} styling mechanism. The latter implicitly requires the \texttt{biblatex} package and the \texttt{biber} backend. It adopts an internal Unicode data model while offloading significantly more logic, specifically sorting, filtering, and formatting, to the macro layer.

Consequently, although both methods "appear to use the same \texttt{.bib} file," they diverge significantly in character encoding support, linguistic processing capabilities, style customization, and compilation steps. A smooth migration cannot be achieved by simply swapping commands. From the user's perspective, this implies that bibliography management is not a stable, interchangeable module, but rather deeply coupled with the engine and packages. Users must not only understand \texttt{.bib} syntax but also explicitly identify which call path their document is entering for bibliography processing. This further exacerbates the cognitive burden regarding tool selection and workflow configuration.

\textbf{Ecosystem Fragmentation:} Certain packages support only specific engines, degrading document portability. A common example is the \texttt{fontspec} package, used for loading system fonts (OpenType/TrueType) and Unicode settings. Under pdfLaTeX, this triggers a fatal error (\texttt{cannot-use-pdftex}) because pdf\TeX{} lacks native Unicode support.

\textbf{Increased Cognitive Load:} These factors compel users to learn not just \LaTeX{} syntax, but also how to select and configure the appropriate compilation engine to achieve their desired document output.

\subsection{Alienation of the Tool Ecosystem}

To compensate for the aforementioned core design deficiencies, the \LaTeX{} ecosystem has spawned a series of complex compensatory mechanisms, which have in turn introduced new problems.

\subsubsection{Multi-pass Compilation: A Fragile Stopgap}

In \LaTeX{}, the accurate generation of cross-references, tables of contents, and bibliographies relies on multiple compilation passes to make up for the absence of internal state. The typical workflow involves a first pass that generates auxiliary files (like \texttt{.aux} and \texttt{.toc}) to record state, followed by subsequent passes that read these files to populate cross-references, page numbers, or chapter titles. If the document includes citations, external tools such as BibTeX or Biber must also be integrated into the process. For example, the standard procedure for a document with references often follows the sequence: \texttt{pdflatex} $\rightarrow$ \texttt{bibtex} $\rightarrow$ \texttt{pdflatex} $\rightarrow$ \texttt{pdflatex}.

This mechanism imposes several burdens:

\begin{itemize}
  \item \textbf{State Fragmentation:} Document state is scattered across multiple external files.
  \item \textbf{Cognitive Load:} Users must master complex ``compilation rituals.'' For instance, failure to run BibTeX results in all citation numbers appearing as \texttt{??}.
  \item \textbf{Debugging Difficulties:} Error tracing is notoriously difficult. An illegal character in a \texttt{.bib} file might trigger an error in the \texttt{\textbackslash bibliography\{refs\}} command, causing the error message to be completely detached from the actual root cause.
\end{itemize}

In extreme cases, cross-references remain generated during the first pass, leaving the user with a mass of undefined numbers or empty tables until the multi-pass cycle is complete. This reliance on external side effects to patch internal architectural flaws is, in essence, a form of technical debt.

\subsubsection{Long-Term Suppression of the Tool Ecosystem}

The batch processing design of \TeX{} has stifled the development of peripheral tools. Real-time preview in \LaTeX{} editors is merely a simulation achieved by frequently triggering background compilations, which suffers from significant latency and layout inconsistencies. For example, in documents containing numerous floating objects and formulas, even modifying a small paragraph can trigger a global re-calculation of pagination, causing the preview to lag. Furthermore, features like intelligent code refactoring or syntax completion are virtually impossible to implement effectively without a structured document tree. Most automation tools rely on regular expression matching rather than syntax tree analysis, leading to limited error-checking capabilities and a high propensity for false positives or missed errors.

This structural dilemma is not accidental; rather, it is rooted in the batch-processing paradigm of \LaTeX{} itself. As quoted in the main text, Frank Mittelbach, the technical lead for \LaTeX{}, candidly admitted in an interview~\cite{interview2021} that the separation between local and global processing ``is right now hard in \TeX{}.'' For completeness, the fuller exchange runs:

\begin{quotation}
  PN: "I try to look at this issue from the point of view of global and local, and interactivity is just like a... This is probably a change that happens very fast, and that you worried only about the local stuff, but the separation between local and global in \LaTeX{} seems to be hard."

  FMi: "First of all, it is right now hard in \TeX{}\ldots"
\end{quotation}

\subsection{Comparison: Design Paradigms of Structured and Incremental Systems}

Systems like GNU \TeX{}macs adopt a distinct architecture: the document maintains a structured tree representation in memory, where local modifications trigger immediate local repainting, and structural validity checks occur during the editing phase. For instance, modifying a formula or a floating object redraws only the affected region. Cross-references and numbering update instantly, ensuring the user always interacts with a predictable and previewable document state. Errors are isolated within local modules, preventing global system crashes.

This contrast shows that \LaTeX{}'s need for multiple compilation rounds and external tools is not unavoidable and instead results from a compensatory design strategy. These limitations could be completely bypassed by refactoring the architecture. Ultimately, the fundamental problem of \LaTeX{} lies not the quality of its typesetting, but the defects of global dependency and state management inherent in its execution model.

\section{Limitations of \TeX{} in User Experience Design}
\label{app:con}
\label{sec:ux}

\TeX{} and its derivative, the \LaTeX{} ecosystem, have long occupied a central position in the field of academic typesetting. However, the overall user experience has failed to evolve in sync with the advancement of computing environments and user expectations. From the continuous bloating of distribution sizes to the performance and interaction barriers imposed by the compilation model, and further to the long-standing engineering defects in language design, the \TeX{} ecosystem falls short in modern writing contexts---particularly regarding "usability," "maintainability," and "adaptability to modern workflows." This section analyzes the limitations of \TeX{}'s user experience design from multiple dimensions, including deployment costs, language structure, and practical user experience.

\subsection{Issues with Distribution Scale and Deployment Models}

\begin{figure}[htbp]
    \centering
    \includegraphics[width=\columnwidth]{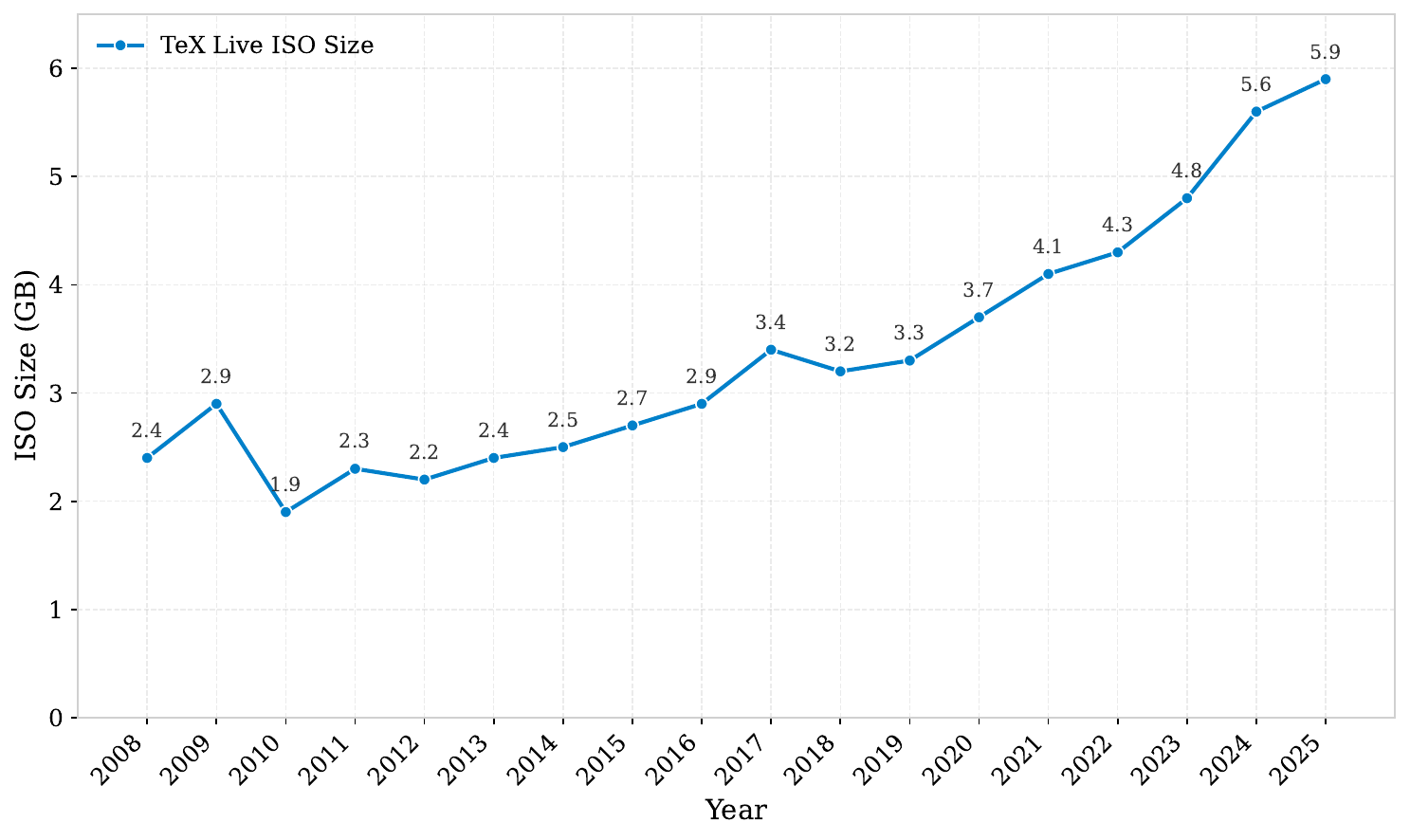}
    \caption{TeX Live ISO Size Trends}
    \label{fig:texlive-size}
\end{figure}

Figure~\ref{fig:texlive-size} illustrates the trend in file size for the most popular TeX Live ISO from 2008 to 2025. The horizontal axis represents the year, and the vertical axis represents file size in GB. Starting at 2.4~GB in 2008, the line shows an overall upward trajectory, reaching 5.9~GB by 2025. Despite minor fluctuations (such as a drop to 1.9~GB in 2010 and a slight dip to 3.2~GB in 2018), the long-term trend demonstrates a nearly twofold increase in size, reflecting the continuous expansion of software packages.

This volumetric growth is primarily driven by TeX Live's role as a comprehensive \LaTeX{} distribution, which constantly integrates new fonts, documentation, multi-language support, and packages to adapt to user needs and technological advancements. For instance, while early versions focused on core functionality, later iterations incorporated extensive PDF support, graphics libraries, and extensions, resulting in volume bloat.

\begin{figure}[htbp]
    \centering
    \includegraphics[width=\columnwidth]{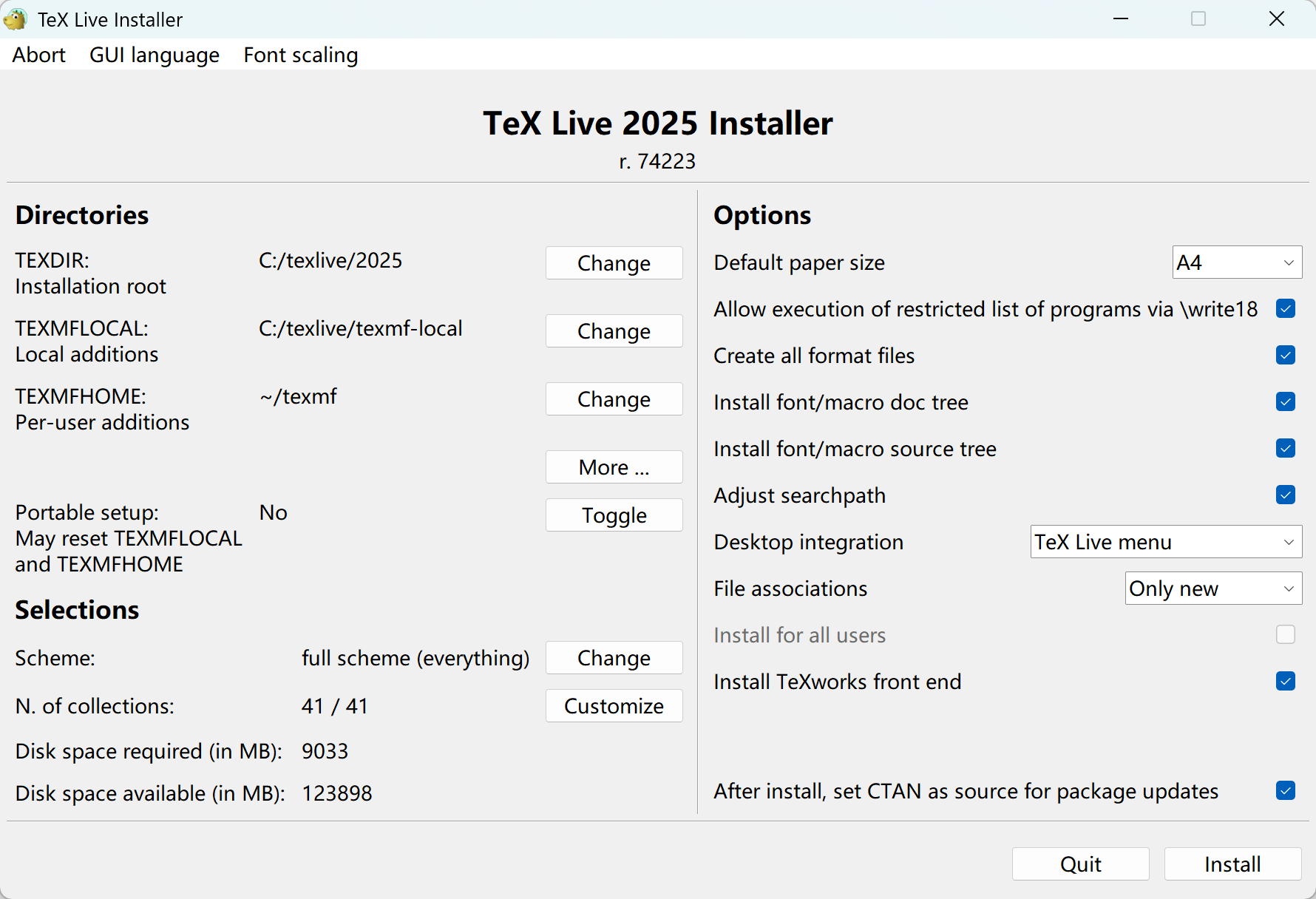}
    \caption{TeX Live Installer}
    \label{fig:texlive-installer}
\end{figure}

However, this continuous expansion warrants critical reflection. Taking TeX Live as an example, a significant portion of its distribution volume consists of documentation sets and font resources. For beginners or light users requiring only basic typesetting functions, such overhead constitutes a clear redundant burden, consuming installation time, disk space, and maintenance effort. Although the \TeX{} ecosystem emphasizes modularity at the package level, the distribution strategy (as illustrated in Figure~\ref{fig:texlive-installer}) still prioritizes full installation as the default. Minimalist installation options, while available, are neither highlighted nor widely adopted. Consequently, most users default to the full version, effectively "normalizing" the issues of bloat and resource waste.

While distributions like MikTeX attempt to address this, a more aggressive promotion of lean versions in default configurations by making documentation, examples, and large font sets optional is far more rational in terms of efficiency, resource utilization, and environmental impact, particularly in network- or storage-constrained scenarios.

Figure~\ref{fig:texlive-installer} illustrates the complex installation interface. In contrast, because the \texttt{.tmu} format encodes document structure explicitly, an editor for \texttt{.tmu} can adopt an on-demand loading mechanism, where features are only loaded when a document structure actually requires them. This significantly reduces the installation size and startup time.

Compounded by its underlying compilation model, \LaTeX{} presents a series of severe usability obstacles. Stemming from inherent contradictions in language design, these issues, manifested as massive installation footprints, slow compilation speeds, and steep learning curves, collectively form a ``deterrent barrier'' for modern users.

\subsection{Real-World Performance Dilemma}

\LaTeX{}'s performance disadvantages warrant discussion, as they render the system incongruous with fast-paced modern workflows. Here, `performance' is not defined by interaction latency or incremental feedback, but rather serves a typesetting model centered on \textbf{batch processing}. The underlying assumption is that the user provides a relatively complete and stable source file, which the system processes through one or more full compilation cycles to generate quality-controlled output. Under this batch-oriented model, optimization prioritizes the correctness and consistency of the final result, rather than immediate responsiveness during the writing process.

However, this throughput-oriented performance goal aligns poorly with modern documentation patterns that prioritize low latency, localized feedback, and continuous interaction. Frequent, minor modifications during writing often trigger a re-parsing and re-typesetting of the entire document, introducing significant wait times that disrupt the continuity between editing and thinking.

Beyond performance unpredictability within a single platform, \LaTeX{} exhibits significant discrepancies in compilation time across different operating systems. For instance, under identical document and package configurations, compilation on Windows is often noticeably slower than on Linux or Unix-like systems. This disparity stems not from hardware differences, but from underlying factors such as file system efficiency, process instantiation overhead, font and I/O management, and the optimization level of the \TeX{} toolchain for specific platforms.

For users, this cross-platform inconsistency undermines system comprehensibility and predictability. The same project may yield vastly different compilation experiences on a personal computer, a lab server, or a cloud environment, making `performance issues' difficult to reproduce, isolate, or optimize. In collaborative scenarios, this variance amplifies into actual collaboration friction, effectively influencing workflow rhythm and tool selection.

Furthermore, \LaTeX{}'s reliance on a multi-stage compilation workflow exacerbates runtime performance costs. Functions such as bibliography management, indexing, and cross-referencing are typically offloaded to multiple external tools, requiring users to run multiple compilation rounds to achieve stable output. This design choice is not driven by performance optimization, but by the language's inherent inability to express complex dependencies within a single compilation pass. Consequently, the overhead of repeated disk I/O, redundant scanning, and process startup is systematically offloaded onto the user.

\subsection{Intrinsic Weakness: Absence of Engineering Standards}

\LaTeXe{} is fundamentally a collection of macros built upon the \TeX{} macro expansion mechanism. Its abstraction capabilities rely primarily on untyped text substitution and grouping scopes, rather than explicit language-level structures. Although the community has actively advanced the \LaTeX3 project in recent years to ameliorate this situation, the project has not been released as a standalone version. Instead, it follows a strategy of progressive evolution, gradually integrating into the existing \LaTeXe{} kernel. Currently, the \texttt{expl3} programming layer provided by \LaTeX3 serves as the foundation for numerous large-scale packages. Consequently, users often utilize its mechanisms indirectly and unconsciously in their daily workflows. This transition primarily results in improved interface consistency, maintainability, and engineering capabilities, accompanied by a degree of performance enhancement.

However, this progressive integration fails to eliminate the long-standing structural contradictions within \LaTeXe{}'s language design. As a system built on a macro expansion language, its approach to modularity, interface expression, and state management stands in fundamental tension with modern engineering practices. Table~\ref{tab:latex-contradictions} systematically outlines these inherent contradictions and their impact on usability and system stability.

\LaTeX{} lacks formal engineering standards for package development. While the \texttt{lppl} license provides some guidance, there are no enforced standards for:

\begin{itemize}
    \item Package documentation quality
    \item API stability and versioning
    \item Error handling and reporting
    \item Testing and validation
\end{itemize}

This has led to a fragmented ecosystem where package quality varies widely, and compatibility issues are common \cite{interview2021}.

\begin{table*}[t]
    \centering
    \small
    \setlength{\tabcolsep}{6pt}
    \caption{Core Structural Contradictions in \LaTeXe{} Language Design and their Engineering Consequences}
    \label{tab:latex-contradictions}
    \begin{tabular}{p{0.18\textwidth}p{0.36\textwidth}p{0.36\textwidth}}
        \toprule
        \textbf{Contradiction}                              & \textbf{Description}                                                                                      & \textbf{Impact}                                                                                                   \\
        \midrule
        Global Naming vs.\ Modularity                       & \LaTeXe{} lacks namespaces; all commands and variables share a global symbol table                        & High risk of package conflicts; loading order directly affects behavior; system fragility                         \\
        Text Substitution vs.\ Structured Interfaces        & Macros are essentially untyped token substitutions, lacking parameter signatures and type systems         & Static checking is difficult; function interfaces are uncomposable; parameter passing is error-prone              \\
        Local State via Grouping vs.\ Compiler Awareness    & State management relies on \TeX{}'s grouping and rollback mechanism rather than explicit variable scoping & Compilers cannot resolve scope; state leakage and logical errors are difficult to detect                          \\
        Conventional Interfaces vs.\ Automatic Verification & Interface constraints rely on documentation and voluntary compliance, not language-level enforcement      & Parameter types/counts cannot be statically verified; errors are only exposed at runtime                          \\
        Compatibility vs.\ Modern Features                  & To maintain compatibility with \LaTeXe{}, implementation of new features is cumbersome                    & Requires reliance on advanced packages like \texttt{xparse} or complex hacks; high learning and maintenance costs \\
        \bottomrule
    \end{tabular}
\end{table*}

In summary, \LaTeXe{} represents a macro language with a severe deficit in engineering rigor. Its system stability, maintainability, and user experience depend heavily on informal conventions and user expertise, rather than structural guarantees provided by the language design itself. This deficiency not only increases the complexity of extension and debugging but also establishes an unavoidable historical burden that constrains future systemic evolution.

\subsection{User Experience Barriers from a Practical Perspective}

The structural contradictions in language design discussed above do not remain abstract concepts. Instead, they directly translate into tangible user experience barriers repeatedly encountered during document composition, package combination, and troubleshooting. These issues do not stem from isolated implementation defects or inadequate documentation, but are intrinsic to the macro expansion mechanism and engineering constraints upon which \LaTeXe{} relies.

\textbf{Lack of Usability and Interface Consistency:} Since command interfaces are constrained by convention rather than language-level mechanisms, \LaTeXe{} lacks a unified standard for parameter syntax, optional argument placement, and starred variants across different packages. A typical example is \verb|\newcommand|, which supports only a single, fixed-position optional argument, whereas more complex interface requirements necessitate the use of packages like \texttt{xparse} (via \verb|\NewDocumentCommand|). This deficiency in interface expressiveness forces users to frequently switch mental models between different packages, significantly increasing the cost of learning and usage.

\textbf{Fragility and Implicit Errors Induced by the Global Namespace:} \LaTeXe{} lacks language-level namespaces and encapsulation mechanisms; all commands and variables share a global symbol table. This design makes it difficult to guarantee independence between packages, with loading order often directly affecting document behavior. In the worst-case scenario, definitions of the same command by different packages may result in ``silent overwriting,'' causing subtle but imperceptible changes in document output, thereby weakening system predictability and reliability.

\textbf{High Opacity in Error Diagnosis and Debugging:} Under an execution model where macro expansion and typesetting decisions are deeply intertwined, error messages are often detached from their root causes. Common alerts like ``\verb|Undefined control sequence|'' or ``\verb|Overfull \hbox|'' typically indicate the surface location where the issue was triggered, rather than the origin of the error. Troubleshooting often forces users to rely on empirical methods like step-by-step commenting and regression testing. This process is both inefficient and difficult to systematize, further exacerbating the maintenance burden for complex documents or large projects.

\textbf{Limited Extensibility and Non-Linear Growth in Implementation Complexity:} When users attempt to implement features with even slight complexity (e.g., multiple optional arguments, starred variants, conditional interfaces, or nested logic), they are often compelled to utilize extensive low-level macro hacks to bypass the language's expressive limitations. Such implementations typically rely on implicit state, special naming conventions, and opaque expansion orders, leading to a significant degradation in code readability and maintainability. Consequently, a linear increase in feature complexity is often accompanied by an exponential rise in implementation complexity.

\textbf{High Learning Costs and the Accumulation of ``Tacit Knowledge'':} Collectively, these issues cause the learning path for \LaTeXe{} to depend heavily on tacit knowledge, including grouping scope rules, internal naming conventions (such as \verb|@|-class commands), macro expansion timing, and execution order. This knowledge is not explicitly expressed through language mechanisms but is scattered across package implementations and community lore, making it difficult for new users to establish a stable, transferable cognitive framework.

\subsection{Contributions and Constraints of \LaTeX3}

The \LaTeX3 project emerged as a response to the structural dilemmas described above. It attempts to introduce a programming layer with distinct software engineering characteristics while preserving the underlying \TeX{} macro expansion model. By systematically standardizing and constraining the macro programming practices of \LaTeXe{} through modular naming conventions, an explicit data type system, and controlled scope management, the project aims to bring order to the system. As its core component, the \texttt{expl3} programming layer improves interface consistency and code maintainability, thereby reducing the risks of naming conflicts and implicit state interference during package integration and extension.

However, the improvements offered by \LaTeX3 remain incremental and limited. Its commitment to long-term backward compatibility with \LaTeXe{} dictates that the new and old programming paradigms coexist for a considerable period. Users are required not only to master the new interface model but also to interact with existing packages and historical conventions. This coexistence complicates the ecosystem and renders the migration process slow and uncertain. It serves as a testament to how the historical baggage accumulated over \LaTeX{}'s long evolution continues to constrain its further engineering development.

\subsection{Limitations of Modern Collaboration Platforms: A Case Study of Overleaf}

In response to \LaTeX{}'s structural deficiencies regarding collaboration support and usability, the academic community has advanced a series of improvement measures. Among these, online collaboration platforms, represented by \textbf{Overleaf}, have emerged as the most influential solution in recent years. By providing an integrated cloud-based environment, Overleaf significantly improves the onboarding experience and multi-user collaboration workflows for \LaTeX{}. However, viewed from the perspective of system architecture, Overleaf's design does not address the fundamental issues of \LaTeX{}'s core model. Its role is closer to the encapsulation and optimization of existing workflows rather than a revolution of the underlying paradigm.

\subsubsection{Key Improvements by Overleaf}

Overleaf has effectively mitigated several obstacles inherent in traditional \LaTeX{} usage through the following mechanisms:

\begin{itemize}
    \item \textbf{Simplified Environment Configuration:} Users are no longer required to install a full \TeX{} distribution (such as TeX Live or MiKTeX) locally. Document composition commences via a web browser, thereby eliminating the technical barrier of local deployment.

    \item \textbf{Automated Compilation Workflow:} The system automatically manages multi-pass compilation and external tool invocations (e.g., BibTeX, MakeIndex, Biber). Users need only trigger a single ``Recompile'' action; the platform's backend manages complex dependencies, reducing the technical knowledge required of the user.

    \item \textbf{Enhanced Collaboration Features:} The platform supports multi-user real-time editing, commenting/annotation, and version history tracking. It provides a foundational experience comparable to modern documentation tools (such as Google Docs), significantly outperforming traditional collaboration methods based on email or Git.
\end{itemize}

These platform-level improvements have significantly expanded the applicability of \LaTeX{} in non-professional typesetting scenarios, such as education and scientific research.

\subsubsection{Constraints of Underlying Architecture}

Despite Overleaf's significant progress in user interface and workflow, its foundation still relies entirely on the traditional \LaTeX{} compilation mechanism~\cite{overleaf_docs}, which introduces several insurmountable limitations:

\begin{itemize}
    \item \textbf{Preview Latency Issues:} The so-called ``real-time preview'' is essentially periodic remote compilation and PDF retrieval, not true instant rendering. When document size increases or complex packages (e.g., \texttt{TikZ}, \texttt{algorithm2e}) are included, compilation time rises significantly, leading to a degraded interactive experience.

    \item \textbf{Limited Conflict Resolution in Collaboration:} Since \LaTeX{} source files are unstructured plain text, the platform cannot parse the document structure at a semantic level. When multiple users simultaneously modify the same paragraph, table, or macro definition, the system can only perform line-based text merging. It is unable to identify logical conflicts, ultimately necessitating manual intervention for resolution.

    \item \textbf{Restricted Functional Extensibility:} Overleaf cannot transcend the boundaries of \LaTeX{}'s own expressive capabilities. For instance, it fails to support true WYSIWYG editing modes, nor does it easily integrate dynamic content (such as interactive charts or real-time data binding). Implementing these features requires a structural overhaul of the typesetting engine, rather than mere adjustments to the frontend interface.
\end{itemize}

\subsubsection{Impact on Trajectory of Technological Evolution}

The success of Overleaf serves as a double-edged sword: while it enhances user experience, it simultaneously masks \LaTeX{}'s inherent technical debt and structural defects, thereby diminishing the motivation to explore next-generation typesetting systems. In contrast, systems such as \TeX{}macs or Typst attempt to refactor the document model and syntax design from the foundation up, yet they struggle to disrupt \LaTeX{}'s dominant position due to entrenched user habits.

Consequently, although Overleaf optimizes the usability of the existing ecosystem and resolves certain surface-level pain points, it objectively establishes a form of ``path dependence,'' delaying the demand for more fundamental technological transformation. This platform-driven optimization renders the legacy system more usable but fails to propel it toward a revolutionary direction characterized by greater structure, extensibility, and interactivity.

\section{Related Work}
\label{app:rew}

The challenges of using \LaTeX{} in LLM-driven workflows have been documented from multiple angles. On the generation side, benchmarks reveal a consistent pattern: LLMs struggle with \LaTeX{}'s long-range syntactic dependencies and implicit semantic conventions. The TeXpert benchmark~\cite{kale2025texpert} reports only 15\% accuracy on complex \LaTeX{} tasks, with logical errors---not superficial typos---accounting for 54\% of failures, demonstrating that even state-of-the-art LLMs struggle with \LaTeX{} code generation. Similar difficulties arise in modality-crossing tasks: \citet{ling_table2latex_2025} found that reinforcement-learning-based table-to-\LaTeX{} conversion suffers from persistent token inefficiency rooted in \LaTeX{}'s verbose syntax, while \citet{xia2024docgenome} performed large-scale benchmarks revealing that Equation-to-\LaTeX{} and Table-to-\LaTeX{} conversion tasks remain challenging even with extensive training data (Edit Distance $> 0.21$), attributing these high edit distances to the information entropy of \LaTeX{} source code. These generation-level failures become systemic when \LaTeX{} is embedded in end-to-end scientific pipelines: \citet{lu2024aiscientist} demonstrated that in a fully automated discovery framework (The AI Scientist), the \LaTeX{} writing stage is a critical bottleneck---the model cannot perceive the rendered PDF, producing overflowing tables and placeholder text that require manual correction. \citet{jain2026bibby} arrived at a complementary conclusion from the editor side: their Bibby AI system found that raw compiler logs are insufficient for error localization and proposed combining logs with a live AST, effectively acknowledging that \LaTeX{}'s batch feedback model must be augmented with structural representations to support reliable AI assistance. \citet{lyn2025translatex} further identify an ``execution illusion'' where LLMs produce linguistically fluent but unexecutable \LaTeX{} code for scientific formatting, introducing TransLaTeX, a reasoning-and-control framework with compiler-level verification. Our work extends this line of reasoning by arguing that such augmentation is inherently limited---a document format designed around explicit tree structure, rather than one retrofitted with AST recovery, can more fundamentally resolve the feedback and semantic transparency problems these systems encounter.

The representation efficiency of \LaTeX{} also poses challenges upstream, during data preparation and model training. \citet{paster2023openwebmath} documented the substantial engineering effort required to extract and normalize 14.7B tokens of mathematical content from web sources containing heterogeneous \LaTeX{} fragments, attaching particular importance to mathematical content for LLM training. \citet{lin2024accurate} showed that direct LLM processing of documents incurs prohibitive costs, achieving up to 30\% reduction only through semantic hierarchical indexing---a workaround necessitated by the verbosity of the underlying format, highlighting the need for more efficient document representations. That markup structure matters for model learning has been demonstrated more directly: \citet{li2022markuplm} showed with MarkupLM that jointly modeling text and markup tags improves document understanding, while \citet{taylor2022galactica} found in training Galactica that \LaTeX{} equations are a first-order component of scientific language whose representational overhead affects model behavior. Taken together, these findings suggest that \LaTeX{}'s syntactic redundancy---where equivalent renderings admit multiple source forms (e.g., \verb|\frac{a}{b}| vs. \verb|{a \over b}|)---reduces next-token predictability and inflates training costs. Evidence from neural code modeling further supports the importance of structured representations: TreeDiff~\cite{zeng2026treediffastguidedcodegeneration} demonstrated that AST-guided span masking significantly outperforms token-level random masking for diffusion-based code generation, particularly for longer sequences (36.59\% vs.\ 33.54\% pass@1 on HumanEval with 1024-token prompts), suggesting that hierarchical representations enable models to better capture long-range dependencies. In Section~\ref{sec:eff-in-sft}, we provide direct experimental evidence for this hypothesis by comparing fine-tuning convergence on \LaTeX{} versus the lower-entropy \texttt{.tmu} format, which provides a more information-dense and structurally explicit representation than linear \texttt{.tex}, enabling more efficient learning.

On the systems side, dissatisfaction with \TeX{}'s batch compilation has produced two distinct architectural responses, neither of which fully addresses the problem our work targets. The first is incremental markup compilation: \citet{madje2023typst} and \citet{haug2022typst} designed Typst with a functional type system and incremental compiler that achieves sub-second preview updates, resolving the latency problem but retaining a source-editing paradigm that still separates authoring from visual output. The second is interactive augmentation of existing \LaTeX{} workflows: \citet{gobert2022ilatex} proposed i-\LaTeX{}, which overlays interactive ``transitional'' widgets onto a code editor to bridge source and rendered output---an approach that alleviates feedback delays without removing the underlying batch compilation dependency. Both strategies leave the core tension unresolved: the document's authoritative representation remains either linear markup (Typst) or \TeX{} source (i-\LaTeX{}), rather than an explicit semantic tree. The structured WYSIWYG approach pioneered by \citet{van_der_hoeven_gnu_2001} with GNU \TeX{}macs took a more radical position, demonstrating that high-quality mathematical typesetting can be achieved within a tree-based editor that maintains structure as its primary representation. Empirical evidence supports the case for such alternatives: \citet{knauff2014efficiency} showed in controlled experiments that \LaTeX{} users can be slower and produce more formatting errors than Word users, while \citet{tan2024inconsistencies} systematically documented visual inconsistencies across \TeX{} engines and TeX Live versions, substantiating the fragility of the ecosystem under real-world version drift. \citet{gardner2025neuralatex} implement a deep learning library entirely in pure \LaTeX{}; when the document is compiled, the \LaTeX{} engine trains the network and generates figures, with their paper taking 48 hours to compile, illustrating the extreme runtime cost of \LaTeX{} when used as a programmable substrate. Our work builds on the \TeX{}macs lineage through the tree-structured \texttt{.tmu} format (Section~\ref{sec:mogan}), and contributes new evidence that this representation not only improves editing responsiveness but also yields measurable advantages for LLM tasks---a dimension absent from prior structured-editor evaluations.

\section{The \texttt{.tmu} format v.s. other editors}
\label{app:mvse}

The \texttt{.tmu} format builds on the document representation of the GNU \TeX{}macs lineage~\cite{van_der_hoeven_gnu_2001, moganstem2025, liiistem2025}, in which the document is stored as an explicit, tree-structured representation rather than as linear markup. This family of formats underpins the WYSIWYG structured editors that maintain document structure as their primary representation. Table~\ref{tab:mogan-vs-other} compares an editor based on the \texttt{.tmu} format with several other editors currently available.

\begin{table}[htbp]
    \centering
    \small
    \setlength{\tabcolsep}{3pt}
    \caption{
        Comparison between a \texttt{.tmu} editor and alternative editors.
        Sources: \texttt{.tmu} format~\cite{moganstem2025},
        \TeX{}macs~\cite{van_der_hoeven_gnu_2001},
        \TeX/\LaTeX~\cite{lamport1994document},
        LyX~\cite{jackson2001lyx},
        Typst~\cite{madje2023typst}.
    }
    \label{tab:mogan-vs-other}
    \begin{tabular}{p{1.9cm}cccc}
        \toprule
        \textbf{Editor}
        & \textbf{WYSI}
        & \textbf{Struct.}
        & \textbf{Unicode}
        & \textbf{\TeX{} Compat.} \\
        \midrule
        \textbf{\texttt{.tmu} (this work)} & Yes              & Yes                 & Yes                      & Partial                       \\
        \textbf{\TeX{}macs}    & Yes              & Yes                 & Partial                  & Partial                       \\
        \textbf{TeX/LaTeX}  & No               & Yes                 & Partial                  & Full                          \\
        \textbf{Word}       & Yes              & No                  & Partial                  & Incompatible                  \\
        \textbf{LyX}        & Partial          & Yes                 & Partial                  & Full                          \\
        \textbf{Typst}      & No               & Yes                 & Yes                      & Partial                       \\
        \bottomrule
    \end{tabular}
\end{table}

The core technical challenge in supporting WYSIWYG editing from a structured representation lies in how mathematical formulas are represented and subsequently rendered. Unlike \TeX{}, which represents formulas through layout-oriented markup in the form of typesetting instructions, the \texttt{.tmu} format adopts a tree-based, functional representation for both mathematical formulas and the document structure itself. The following discussion evaluates this representation with specific examples.

\section{Fast Reference Rendering}
\label{app:ren}

Taking bibliographic citations as an instance: in the \LaTeX{} ecosystem, \verb|\cite| serves merely as a macro interface, with actual citation relationships established indirectly via intermediate files such as \verb|.aux| and \verb|.bbl|. Consequently, any modification to bibliography entries typically necessitates multiple global compilation cycles, exemplifying a classic batch processing model. In contrast, the \texttt{.tmu} format encodes citation relationships as structural links directly embedded in the document tree. To update a reference, a structured editor for \texttt{.tmu} performs only a local search and re-renders the specific leaf nodes involved, thereby avoiding a full re-parsing of the document.

\section{On-demand plugin loading}
\label{app:plug}

Because the \texttt{.tmu} format encodes document structure explicitly, a \texttt{.tmu} editor can adopt a monolithic installation paradigm, diverging from the package repository distribution model exemplified by TeX Live. A single installation package delivers a complete and self-contained editing and typesetting system, encompassing core executables, a built-in scripting runtime, a document model, and a rendering engine. Consequently, users avoid managing numerous independent packages or resolving granular macro dependencies. Although functional extensions exist as plugins, they do not load upon startup; instead, they load dynamically at runtime only when specific document structures in the \texttt{.tmu} tree trigger the corresponding requirements. This on-demand mechanism ensures that complexity is dictated by document content rather than a pre-configured feature set.

It is noteworthy that a \texttt{.tmu} editor requires an installation package of merely one hundred megabytes, occupying only several hundred megabytes of local storage upon extraction, yet it offers immediate, out-of-the-box support for the editing and export of complex mathematical documents. In stark contrast, even a minimalist \LaTeX{} installation typically demands a local environment exceeding 1 GB, while a full distribution reaches approximately 6 GB in 2025.

This disparity stems from a fundamental divergence in system design. \LaTeX{} front-loads "potential future capabilities" as an installation cost, maintaining compatibility through its macro language and package ecosystem. Conversely, the \texttt{.tmu} approach defers complexity to runtime, confining it within the actual execution path. Extensibility is achieved via a unified document tree model and a runtime plugin mechanism. Consequently, complexity is activated dynamically by specific documents at runtime, whereas \LaTeX{}'s complexity manifests primarily during the installation and configuration phases.

In summary, TeX Live installs an ever-accumulating repository of historical packages, whereas the \texttt{.tmu} approach provides an evolvable document system. The complexity of the former is front-loaded to the installation phase, while the latter is triggered by the document at runtime.

\section{Benchmark on compiling/rendering time}
\label{app:bench}

\begin{figure}[htbp]
    \centering
    \includegraphics[width=\columnwidth]{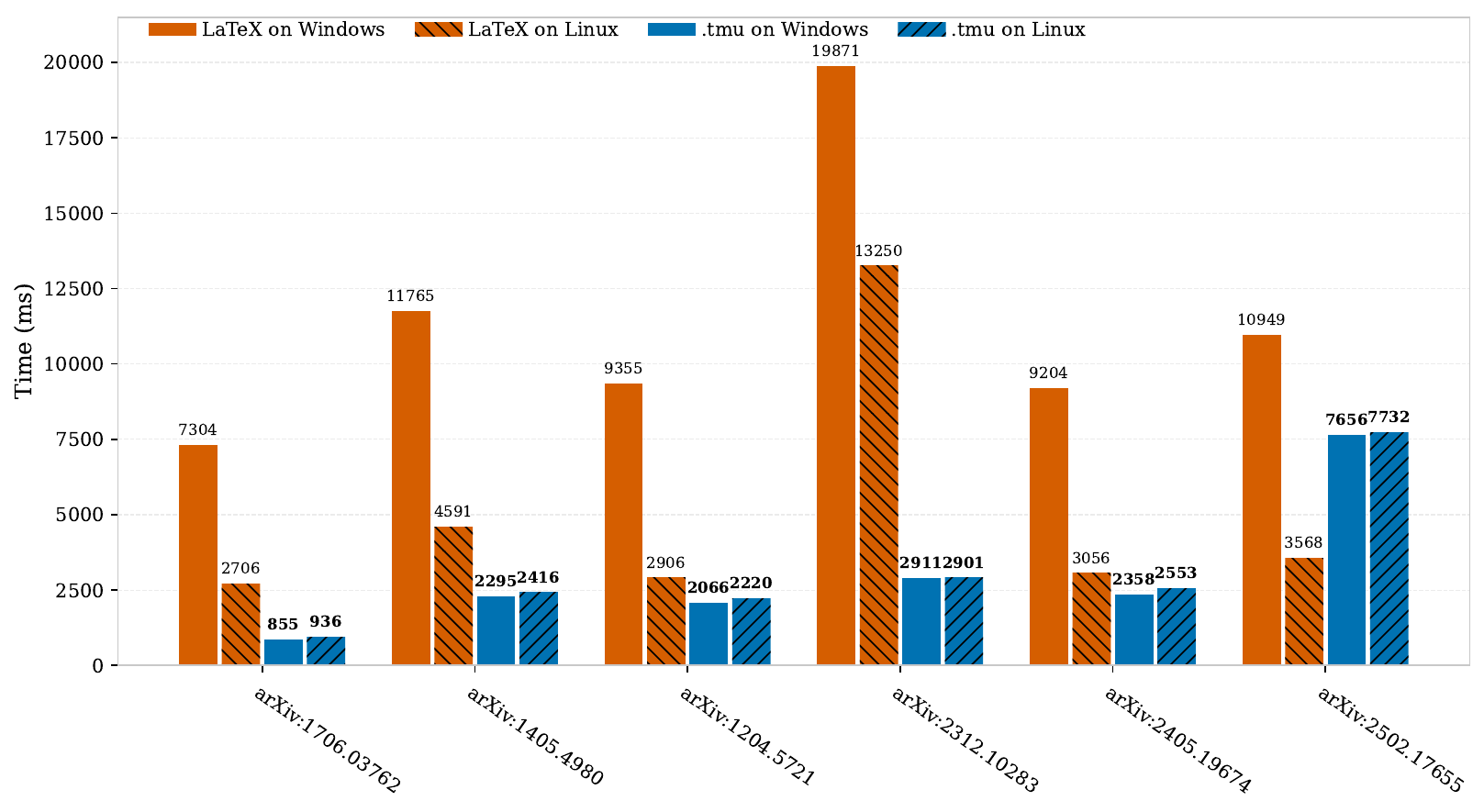}
    \caption{Benchmark on full compilation. Each bar represents the average of three trials.}
    \label{fig:full-compile}
\end{figure}

Limited by the design of \LaTeX{}, the compilation process requires significant time for documents that are rich in cross-references, tables of contents, and bibliographies. We chose 6 papers from arXiv that satisfy this richness criterion as benchmark documents. Note that a \texttt{.tmu} editor is a WYSIWYG editor, so the comparison is unfair! The time consumption for the \texttt{.tmu} editor is
\[
    t_{\text{compiling}} + t_{\text{rendering}} + t_{\text{IO}},
\]
where $t_{\text{compiling}}$ is the compiling time, $t_{\text{rendering}}$ is the rendering time, and $t_{\text{IO}}$ is the \emph{extra} I/O overhead for WYSIWYG editing. As shown in Figure~\ref{fig:full-compile}, even with the extra I/O process, the \texttt{.tmu} editor outperforms \LaTeX{} in compiling/rendering time for most documents. Note that for document \texttt{arXiv:2502.17655}, the \texttt{.tmu} editor compiles and renders slower than \LaTeX{}; this exception is due to the size of the document, as it has 120 pages. Therefore, the I/O overhead accounts for a large proportion.

Another limitation of \LaTeX{} is slow incremental updates. The incremental-update experiment (adding new sections, adding tables in some paragraphs, modifying the relations between labels and references, and rearranging the positions of content slightly) is summarized in Section~\ref{sec:experiments}. Note that for document \texttt{arXiv:2502.17655}, the \texttt{.tmu} editor outperforms \LaTeX{} on incremental updates, which seems to contradict the result of the full-compilation experiment above. The reason for such ``contradiction'' is due to the I/O overhead: with the \texttt{.tmu} format, the I/O process is only activated when the user opens the document. Therefore, for incremental updates, $t_{\text{IO}}=0$.

\section{Prompts for evaluating structure locating}
\label{app:20ques}

You are an expert in \LaTeX{}. Your task is to read the \texttt{main.tex} and answer the following questions:

\begin{enumerate}
    \item Count the number of sections.
    \item Count the number of subsections.
    \item Count the number of figures and tables.
    \item Count the number of cross-references and bibliography references.
    \item In which section does Equation 10.15 appear?
    \item In which subsection does Equation 8.66 appear?
    \item Is there a direct proof below Equation 12.1?
    \item In what context does formula A.3 appear?
    \item In which environment is Definition 4.4 first cited?
    \item What is the number of the first equation after the first citation of Definition 4.4?
    \item In which environment is Definition 7.1 first cited?
    \item What is the number of the first equation after the first citation of Definition 7.1?
    \item How many steps are there in the proof of Lemma 6.4?
    \item Which definition or lemma numbers are directly used in the proof of Lemma 6.4?
    \item How many steps are there in the proof of Lemma 8.3?
    \item Which definition or lemma numbers are directly used in the proof of Lemma 8.3?
    \item In which section does Citation 1 first appear?
    \item In which subsection does Citation 6 first appear?
    \item What was Citation 25 originally used to prove?
    \item Has Citation 31 appeared in the article?
\end{enumerate}

\end{document}